# Consistency and Random Constraint Satisfaction Models

**Yong Gao**                                                    YONG.GAO@UBC.CA
*Irving K. Barber School of Arts and Sciences*
*University of British Columbia Okanagan, Kelowna, Canada V1V 1V7*
**Joseph Culberson**                                            JOE@CS.UALBERTA.CA
*Department of Computing Science, University of Alberta*
*Edmonton, Alberta, Canada T6G 2E8*

## Abstract

In this paper, we study the possibility of designing non-trivial random CSP models by exploiting the intrinsic connection between structures and typical-case hardness. We show that constraint consistency, a notion that has been developed to improve the efficiency of CSP algorithms, is in fact the key to the design of random CSP models that have interesting phase transition behavior and guaranteed exponential resolution complexity without putting much restriction on the parameter of constraint tightness or the domain size of the problem. We propose a very flexible framework for constructing problem instances with interesting behavior and develop a variety of concrete methods to construct specific random CSP models that enforce different levels of constraint consistency.

A series of experimental studies with interesting observations are carried out to illustrate the effectiveness of introducing structural elements in random instances, to verify the robustness of our proposal, and to investigate features of some specific models based on our framework that are highly related to the behavior of backtracking search algorithms.

## 1. Introduction

The tale of random models of constraint satisfaction problems (CSP) gives us an excellent example of the dramatic impact of structures on the typical-case hardness of randomly-generated problem instances (Achlioptas et al., 1997; Gent et al., 2001; MacIntyre et al., 1998; Molloy, 2002; Prosser, 1996; Smith, 2001). It also shows that the study of phase transitions of random models of NP-complete problems is indeed a viable approach to understanding the interplay between the structures and the efficiency of search algorithms.

Unlike the random Boolean satisfiability problem (SAT) where structures do not seem to exist in randomly-generated instances, algorithmically-exploitable structures do exist in the more general random constraint satisfaction problem. The chief reason for this is that in random constraint satisfaction instances the uncontrolled existence of multiple nogoods within the same constraint generates small scale structures that make the instance unsatisfiable. Here small scale may mean only a few variables are involved, or that the variables involved are very highly constrained in a local way. The first example of such structures is the existence of flawed variables and flawed constraints in random instances generated from the four classical random CSP models in most of the parameter regions (Achlioptas et al., 1997), assuming that the domain size is fixed. The appearance of flawed variables and constraints makes these models trivially unsatisfiable and excludes any phase transition behavior. As yet another example, the existence of embedded easy subproblems (Gao & Culberson, 2003), also called "flowers" (Molloy & Salavatipour, 2003), in another part of the parameter space of the models makes randomly-generated instances polynomially sovable by constraint consistency algorithms.

Several new models have been proposed to overcome the trivial unsatisfiability. Gent et al. (2001) proposed the flawless random binary CSP based on the notion of a *flawless conflict matrix*. Instances of the flawless random CSP model are guaranteed to be arc-consistent, and thus do not suffer asymptotically from the problem of flawed variables. Achlioptas et al. (1997) proposed a nogoods-based CSP model and showed that the model has non-trivial asymptotic behaviors. Random CSP models with a (slowly) increasing domain size have also been shown to be free from the problem of flawed variables and have interesting threshold behavior (Xu & Li, 2000; Smith, 2001; Frieze & Molloy, 2003).

While all of these proposed models are guaranteed to have interesting phase transition behavior (some of them are also guaranteed to generate hard instances at phase transitions), the fundamental relation between the structures and the typical-case hardness of randomly-generated CSP instances has not been seriously addressed. As a consequence, superficial conditions and restrictions on the parameter of constraint tightness in these models frequently give the (false) impression that it is the constraint tightness and/or the domain size that determines the behavior of random CSP instances. For the classic binary random CSP, the constraint tightness has to be less than the domain size in order to have a phase transition. The flawless random CSP does have a true solubility phase transition at a high constraint tightness, but as we will show later, it still suffers from embedded easy unsatisfiable subproblems when the constraint tightness is greater than the domain size. In CSP models with increasing domain size, there is still an obvious restriction on the possible values of the constraint tightness. In the nogood-based CSP model, it is impossible to have a high constraint tightness without making the constraint (hyper)graph very dense.

In this paper, we study the possibility of designing non-trivial random CSP models by exploiting the intrinsic connection between structures and typical-case hardness. For this purpose, we show that *consistency*, a notion that has been developed to improve the efficiency of CSP algorithms, is in fact the key to the design of random CSP models that have interesting phase transition behaviors and guaranteed exponential resolution complexity without putting much restriction on the parameter of constraint tightness or the domain size.

In Section 4 we report a series of experimental studies. Algorithmic studies on random instances are sometimes criticized for being more about the model than about the algorithms (Johnson, 2002). We believe that it is better to study the interaction of models with algorithms. For example, in Section 4.2 we observe a double peak phenomenon where our best guess suggests it is related to the (lack of) consistency enforcement in the algorithms we use interacting with the degree of consistency enforced in the instances by the various models. We propose a very flexible framework for constructing problem instances with interesting behavior and develop a variety of concrete methods to construct specific random CSP models that enforce different levels of constraint consistency. We hope this framework will be useful in helping researchers construct pseudo-random instances exhibiting various kinds of structure, while avoiding trivialities that have sometimes tainted claimed results in the past, or that may cause algorithm design to become too focused on a particular structure to the detriment of a more general application.

## 2. Random Models for CSPs

Throughout this paper, we consider binary CSPs defined on a domain $D$ with $|D| = d$. A binary CSP $\mathcal{C}$ consists of a set of variables $x = \{x_1, \cdots, x_n\}$ and a set of binary *constraints* $\{C_1, \cdots, C_m\}$. Each constraint $C_i$ is specified by its *constraint scope*, a pair of variables in $x$, and a *constraint relation* $R_{C_i}$ that defines a set of incompatible value-tuples in $D \times D$ for the scope variables. An incompatible value-tuple is also called a *restriction*, or a *nogood*. Associated with a binary CSP is a *constraint graph* whose vertices correspond to the set of variables and whose edges correspond to the set of constraint scopes. Throughout the discussion, we assume that the domain size is a fixed constant. In the rest of the paper, we will be using the following notation:

1. $n$, the number of variables;

2. $m$, the number of constraints;

3. $d$, the domain size; and

4. $t$, the constraint tightness, i.e., the size of the restriction set.

Given two variables, their constraint relation can be specified by a 0-1 matrix, called the *conflict matrix*, where an entry 0 at $(i, j)$ indicates that the tuple $(i, j) \in D \times D$ is incompatible. Another way to describe a constraint relation is to use the *compatibility graph*, a bipartite graph with the domain of each variable as an independent part, where an edge signifies the corresponding value-tuple is compatible.

We now define our generic model for generating random binary CSPs. Recall that a relation is a set of ordered pairs.

**Definition 2.1. ($\mathcal{B}_{n,m}^{d,t}[\mathcal{L}]$, Restricted Random Binary CSP)** *Let $d, t, n, m$ be integers as specified above, all variables share a common domain $D = \{1, \ldots, d\}$, and let $\mathcal{L} = \{L_1, L_2, \ldots \mid L_i \subset D \times D\}$ be a set of relations. $\mathcal{B}_{n,m}^{d,t}[\mathcal{L}]$ is a random CSP model such that*

1. *its constraint graph is the standard random graph $G(n, m)$ where the $m$ edges of the graph are selected uniformly at random from all possible $\binom{n}{2}$ edges; and*

2. *for each edge of $G(n, m)$, a constraint relation on the corresponding scope variables is specified as follows:*

   (a) *Choose at random a relation $L \in \mathcal{L}$.*

   (b) *Choose $t$ value-tuples from $D \times D \setminus L$ uniformly at random as its nogood set.*

To the above definition, we typically want to add the condition that the set $\mathcal{L}$ satisfies

$$(a, a) \notin \bigcap_{i \geq 1} L_i, \forall a \in D.$$

If this condition is not met, then the instances generated by the CSP model are always trivially satisfiable (and vacuously have exponential resolution complexity). This is a very mild, but sufficient, condition to make sure that $\mathcal{B}_{n,m}^{d,t}[\mathcal{L}]$ has a linear solubility threshold. In particular, all the existing models and all the models to be discussed in the present paper satisfy the condition. In Appendix A.3, we provide a result showing that under this condition, there is a constant $c^*$ such that $\mathcal{B}_{n,m}^{d,t}[\mathcal{L}]$ with $m/n > c^*$ is asymptotically unsatisfiable with probability one.

By placing other restrictions on $\mathcal{L}$ we can subsume many of the previously existing models as well as the new models we will define shortly.

**Definition 2.2. (Model B: Random Binary CSP $\mathcal{B}_{n,m}^{d,t}$)** *If for $\mathcal{B}_{n,m}^{d,t}[\mathcal{L}]$ we have $\mathcal{L} = \{\emptyset\}$, we write $\mathcal{B}_{n,m}^{d,t}$, which is known in the literature as Model B.*

An instance of a CSP is said to be $k$-*consistent* if and only if for any $(k-1)$ variables, each consistent $(k-1)$-tuple assignment to the $(k-1)$ variables can be extended to a consistent $k$-tuple assignment to any other $k$th variable. A CSP instance is called *strongly $k$-consistent* if and only if it is $j$-consistent for each $j \leq k$. Of special interest are the strong-k-consistencies for $k = 1, 2, 3$, also known as *node-consistency, arc-consistency, and path-consistency* (Mackworth, 1977).

Note that in $\mathcal{B}_{n,m}^{d,t}$ there are no restrictions on the construction of the constraint. It has been shown that $\mathcal{B}_{n,m}^{d,t}$ is asymptotically trivial and unsatisfiable for $t \geq d$, but has a phase transition in satisfiability probability for $t < d$ (Achlioptas et al., 1997; Gent et al., 2001). Basically, the trivial cases arise because asymptotically with high probability there will be a variable such that all of its values are forbidden due to constraints containing it that are not arc-consistent. This motivates the introduction of the flawless conflict matrix to make sure that the random model is arc-consistent (Gent et al., 2001).

**Definition 2.3. (Flawless Random Binary CSP $\mathcal{B}_{n,m}^{d,t}[\mathcal{M}]$)** *If for $\mathcal{B}_{n,m}^{d,t}[\mathcal{L}]$ each $L \in \mathcal{L}$ is a bijection, $L : D \leftrightarrow D$, then we write $\mathcal{B}_{n,m}^{d,t}[\mathcal{M}]$. This is the flawless random binary CSP as defined by Gent et al. (2001).*

The $\mathcal{M}$ in the notation is a reference to matchings. If we use the construction for the generalized flawless random binary CSPs that we present in Section 3.2, but restrict the underlying graphs to be matchings, then it reduces to the flawless model.

By specifying bijections we create a set of tuples such that there is one for each $i \in D$. These will not be considered when choosing incompatible value-tuples, and so the resulting model is guaranteed to be arc-consistent and consequently will not have flawed variables. However, even though the flawless random binary CSP $\mathcal{B}_{n,m}^{d,t}[\mathcal{M}]$ does not suffer from the problem of trivial unsatisfiability, it can be shown that $\mathcal{B}_{n,m}^{d,t}[\mathcal{M}]$ may asymptotically have embedded easy subproblems for $t \geq d - 1$.

**Theorem 2.1.** *For $t \geq d-1$, there is a constant $c^* > 0$ such that for any $\frac{m}{n} > c^*$, with high probability $\mathcal{B}_{n,m}^{d,t}[\mathcal{M}]$ is asymptotically unsatisfiable and can be solved in polynomial time.*

*Proof.* We outline the proof here and provide a detailed proof in Appendix A.1. The idea is to show that for sufficiently high $\frac{m}{n}$, $\mathcal{B}_{n,m}^{d,t}[\mathcal{M}]$ contains structures (sub-instances) that (1) cause the unsatisfiability of the whole instance and (2) can be detected in polynomial time.

One such structure is the so-called "flower" consisting of a collection of forbidden constraint cycles which we explain below.

A binary constraint over two variables $x_1$ and $x_2$ is said to contain an $(\alpha, \beta)$-forcer if for the assignment $x_1 = \alpha$, the only compatible assignment to $x_2$ is $x_2 = \beta$. Consider a CSP instance. Let $\{x_0, x_1, \cdots, x_{r-1}\}$ be a subset of $r$ variables and $\alpha$ be a domain value. Suppose that there is a cycle of constraints

$$C_1(x_0, x_1), C_2(x_1, x_2), \ldots, C_{r-1}(x_{r-2}, x_{r-1}), \text{ and } C_r(x_{r-1}, x_0)$$

such that $C_1(x_0, x_1)$ contains an $(\alpha, \alpha_1)$-forcer, $C_i(x_{i-1}, x_i)$ contains an $(\alpha_{i-1}, \alpha_i)$-forcer, and $C_r(x_{r-1}, x_0)$ contains an $(\alpha_{r-1}, \beta)$-forcer with $\alpha \neq \beta$. Then, the assignment $x_0 = \alpha$ cannot be used in any satisfying assignment. We call such a cycle of constraints a forbidden cycle (for the variable $x_0$ and domain value $\alpha$). Now, if there are $d$ (the domain size) forbidden cycles for the variable $x_0$, one for each domain value, then the CSP instance is not satisfiable since none of its domain values can be in a satisfying assignment. We call a variable together with forbidden cycles for each $\alpha$ a flower.

Using the second-moment method, it can be shown that there is a constant $c^*$ such that for any $\frac{m}{n} > c^*$, the probability that $\mathcal{B}_{n,m}^{d,t}[\mathcal{M}]$ contains a flower is asymptotically one. So, with high probability, $\mathcal{B}_{n,m}^{d,t}[\mathcal{M}]$ is unsatisfiable if $\frac{m}{n} > c^*$. Furthermore, if a binary CSP instance contains a flower, then any path-consistency algorithm (e.g., Mackworth, 1977) will produce a new CSP instance in which the variable of the flower has an empty domain. Therefore, a CSP instance with an embedded flower sub-instance can be solved in polynomial time. $\qquad\square$

It should be noted that $\mathcal{B}_{n,m}^{d,t}[\mathcal{M}]$ does have a non-trivial phase transition since it is satisfiable with high probability if $\frac{m}{n} < \frac{1}{2}$. Theorem 2.1 does not exclude the possibility that $\mathcal{B}_{n,m}^{d,t}[\mathcal{M}]$ will be able to generate hard instances when $\frac{m}{n}$ is below the upper bound $c^*$, in particular in the case of a large domain size. Further investigation is required to fully understand the complexity of $\mathcal{B}_{n,m}^{d,t}[\mathcal{M}]$ in this regard.

## 3. Consistency, Resolution Complexity, and Better Random CSP Models

Propositional resolution complexity deals with the minimum length of resolution proofs for an (unsatisfiable) CNF formula. As many backtracking-style complete algorithms can be simulated by a resolution proof, the resolution complexity provides an immediate lower bound on the running time of these algorithms. Since the work of Chvatal and Szemeredi (1988), there have been many studies on the resolution complexity of randomly-generated CNF formulas (Beame et al., 1998; Achlioptas et al., 2001).

Mitchell (2002b) developed a framework in which the notion of resolution complexity is generalized to CSPs and the resolution complexity of randomly-generated random CSPs can be studied. In this framework, the resolution complexity of a CSP instance is defined to be the resolution complexity of a natural CNF encoding which we give below. Given an instance of a CSP on a set of variables $\{x_1, \cdots, x_n\}$ with a domain $D = \{1, 2, \cdots, d\}$, its CNF encoding is constructed as follows:

1. For each variable $x_i$, there are $d$ Boolean variables $x_i : 1, x_i : 2, \ldots, x_i : d$ each of which indicates whether or not $x_i$ takes on the corresponding domain value. There is a clause $x_i : 1 \lor x_i : 2 \lor \ldots \lor x_i : d$ on these $d$ Boolean variables making sure that $x_i$ takes at least one of the domain values;

2. For each restriction $(\alpha_1, \cdots, \alpha_k) \in D^k$ of each constraint $C(x_{i_1}, \cdots, x_{i_k})$, there is a clause $\overline{x_{i_1} : \alpha_1} \lor \cdots \lor \overline{x_{i_k} : \alpha_k}$ to respect the restriction.

This CNF encoding is equivalent to the original CSP problem in the sense that a CSP instance is satisfiable if and only if the CNF encoding is satisfiable. Any satisfying assignment to the CSP variables translates immediately to a satisfying assignment of the CNF encoding. On the other hand, any satisfying assignment to the CNF encoding can also be translated to a satisfying assignment to the original CSP instance. If in the CNF assignment, more than one of the $d$ boolean variables $x_i : 1, x_i : 2, \ldots x_i : d$ are assigned TRUE, we can just pick any one of them and assign the corresponding domain value to the CSP variable $x_i$. It is possible to add another set of 2-clauses for each constraint variable $x_i$ of the form $\{\overline{x_i : \alpha_j} \lor \overline{x_i : \alpha_h} \mid 1 \le j < h \le d\}$ to ensure that any satisfying assignment to the CNF encoding assigns a unique domain value to each CSP variable. However, these do not in general change the complexity results, and make the analysis more complicated.

Mitchell (2002b) as well as Molloy and Salavatipour (2003) showed that random CSPs will have an exponential resolution complexity if their constraint tightness $t$ is less than a certain value. For random binary CSP $\mathcal{B}_{n,m}^{d,t}$, the requirement is (1) $t < d - 1$; or (2) $t < d$ and $\frac{m}{n}$ is sufficiently small. For $t \ge d - 1$, recent theoretical results (Gao & Culberson, 2003; Molloy & Salavatipour, 2003) indicate that it is still possible for these classical models to have an asymptotic polynomial complexity due to the existence of embedded easy subproblems.

In the following, we will show that it is not necessary to put restrictions on the constraint tightness in order to have a guaranteed exponential resolution complexity. Based on similar arguments as those in the literature (Mitchell, 2002b; Molloy & Salavatipour, 2003; Beame, Culberson, Mitchell, & Moore, 2005), it can be shown that if in $\mathcal{B}_{n,m}^{d,t}[\mathcal{L}]$, the constraint relation of each constraint is chosen in such a way that the resulting instances are always strongly 3-consistent, then $\mathcal{B}_{n,m}^{d,t}[\mathcal{L}]$ has an exponential resolution complexity no matter how large the constraint tightness is.

**Theorem 3.1.** *Let $\mathcal{L}$ be such that $\mathcal{B}_{n,m}^{d,t}[\mathcal{L}]$ is strongly 3-consistent. Then for any constant $\frac{m}{n} = c > 0$, the resolution complexity of $\mathcal{B}_{n,m}^{d,t}[\mathcal{L}]$ is almost surely exponential.*

*Proof.* See Appendix A.2. □

Using the tool developed by Molloy and Salavatipour (2003), the requirement that CSP instances be strongly 3-consistent to have an exponential resolution complexity can be further relaxed. Recall that a constraint on $x_1, x_2$ contains an $(\alpha, \beta)$-*forcer* if for $x_1 = \alpha$ the only consistent assignment allowed by the constraint is $x_2 = \beta$.

**Definition 3.1.** *We call a CSP instance **weakly** 4-**consistent** if*

*1. it is arc-consistent,*

*2. it contains no forcer, and*

3. *for any 4 distinct variables $x_1, x_2, x_3, x_4$ and constraints $C_1(x_1, x_2), C_2(x_2, x_3), C_3(x_3, x_4)$ and for any $\alpha_1, \alpha_4 \in D$ there exist $\alpha_2, \alpha_3 \in D$ such that the assignment $x_1 = \alpha_1, x_2 = \alpha_2, x_3 = \alpha_3, x_4 = \alpha_4$ is consistent with $C_1, C_2$ and $C_3$.*

Note that weak 4-consistency is weaker than strong 3-consistency since it does not require an instance to be 3-consistent.

**Theorem 3.2.** *Let $\mathcal{L}$ be such that $\mathcal{B}_{n,m}^{d,t}[\mathcal{L}]$ is weakly 4-consistent. Then for any constant $\frac{m}{n} = c > 0$, the resolution complexity of $\mathcal{B}_{n,m}^{d,t}[\mathcal{L}]$ is almost surely exponential.*

*Proof.* See Appendix A.2. □

The proofs of the above theorems are based on the structural properties of the underlying constraint random graph of the CSP models and the relation between the size of a resolution proof and the maximum size of a clause in the resolution proof. Below we briefly discuss the key ideas of the proof of Theorem 3.1 to illustrate the role of constraint consistency. First, in order for a CSP instance to have an exponential size proof, the size of a minimum unsatisfiable sub-instance (MUS) must not be constant. Otherwise, an enumeration of all the partial assignments to all the sub-instances of size less than that of the MUS gives us a proof of polynomial size. In the case of $\mathcal{B}_{n,m}^{d,t}[\mathcal{L}]$, the minimum vertex degree of a MUS must be larger than 2 since it is 3-consistent. Due to the local sparseness of a random graph forced by the assumption $\frac{m}{n} = c$, the minimum size of a MUS has to be linear in the problem size $n$.

Second, Ben-Sasson and Wigderson (2001) have shown that to establish an exponential lower bound on the resolution complexity, it is sufficient to show that any resolution proof must contain a clause whose size is linear in the problem size. Let $s$ be the minimum size of a MUS. It can be shown that any resolution proof must contain a clause C such that the minimum size of the sub-instance $\mathcal{J}$ that implies $C$ is at least $\frac{s}{2}$. Since the instance is 3 consistent and $\mathcal{J}$ is minimal, the clause $C$ must contain one literal related to each of the variables in $\mathcal{J}$ whose vertex degree is less than 3. A random graph argument shows that the number of such variables in any sub-instance is linear in $n$. In summary, due to constraint consistency and the local sparseness of a random constraint graph, any resolution proof must contain a clause whose derivation involves a linear number of variables.

The question remaining to be answered is whether or not there are any natural random CSP models that are guaranteed to be strongly 3-consistent or weakly 4-consistent. In fact, the CSP-encoding of random graph $k$-coloring is strongly $k$-consistent. In the rest of this section, we discuss how to generate random CSPs with a high tightness that are strongly 3-consistent or weakly 4-consistent.

### 3.1 The Generalized Flawless Random Binary CSP

We now introduce our generalized flawless model. We call this a generalized flawless model since it requires even stronger conditions on $\mathcal{L}$ than the flawless model of Gent et al. (2001). We assume a common domain $D = \{1, \ldots, d\}, d > 2$.

**Definition 3.2.** *We say that $\mathcal{L}$ is SC-inducing if*

1. *$\forall L \in \mathcal{L}$ and $\forall \alpha \in D$ there exist $\beta, \gamma \in D$ such that $(\alpha, \beta) \in L$ and $(\gamma, \alpha) \in L$ and*

2. $\forall L_1, L_2 \in \mathcal{L}$ and $\forall \alpha, \gamma \in D$ there exists $\beta \in D$ such that $(\alpha, \beta) \in L_1$ and $(\beta, \gamma) \in L_2$.

The first condition in Definition 3.2 guarantees arc consistency in any constraint. The second condition ensures 3-consistency. Given $d > 2$, these are the two conditions for strong 3-consistency.

**Definition 3.3.** *We say that $\mathcal{L}$ is* WC-inducing *if*

1. $\forall L \in \mathcal{L}$ and $\forall \alpha \in D$ there exist $\beta_1, \beta_2, \gamma_1, \gamma_2 \in D$ such that $(\alpha, \beta_1), (\alpha, \beta_2) \in L$ and $(\gamma_1, \alpha), (\gamma_2, \alpha) \in L$ and

2. $\forall L_1, L_2, L_3 \in \mathcal{L}$ and $\forall \alpha, \delta \in D$ there exist $\beta, \gamma \in D$ such that $(\alpha, \beta) \in L_1, (\beta, \gamma) \in L_2$ and $(\gamma, \delta) \in L_3$.

The first condition in Definition 3.3 is to prevent forcers as well as to enforce arc consistency, which are the first two conditions required by the definition of weak 4-consistency.

**Definition 3.4. (Generalized Flawless Random Binary CSPs)**
*For the model $\mathcal{B}_{n,m}^{d,t}[\mathcal{L}]$ if $\mathcal{L}$ is SC-inducing then we write $\mathcal{B}_{n,m}^{d,t}[SC]$. If $\mathcal{L}$ is WC-inducing then we write $\mathcal{B}_{n,m}^{d,t}[WC]$. In either case, we say we generate a generalized flawless random binary CSP.*

**Lemma 3.3.** *A CSP constructed by $\mathcal{B}_{n,m}^{d,t}[SC]$ is strongly 3-consistent. A CSP constructed by $\mathcal{B}_{n,m}^{d,t}[WC]$ is weakly 4-consistent.*

We should note that just as the flawless method could not generate every arc consistent CSP (Gent et al., 2001) not every SC (WC) CSP will be generated by our system. For example, if there is a constraint $C$ on variables $x_1$ and $x_2$ then with respect to a third variable $x_3$ strong 3-consistency requires only that there be a consistent assignment to $x_3$ for those pairs of values that are compatible in $C$. Our method ensures consistency on every pair of values for $x_1, x_2$ independently of whether or not there is a constraint on them. We see no reasonable way to produce the relaxed version given that the constraints have to be generated independently of $G(n, m)$.

### 3.2 Constructing Consistent-Inducing Sets $\mathcal{L}$

We now provide methods to generate a variety of sets of relations $\mathcal{L}$ which are SC-inducing or WC-inducing. Although our CSP models require all variables to have a common domain, we will start with a more general construction on varying domain sizes. This will enable us to significantly increase the domain tightness on larger domains. As a side benefit, it indicates how the models might be adapted to varying domains. However, to build a model for varying domain sizes would require at least that we have distinct sets $L$ for every ordered pair of distinct domains. It would also complicate the selection step, 2(a) in Definition 2.1 as well as the analysis. We leave further development of such models for future research.

We start by constructing a set of bipartite graphs $\mathcal{G}$, which we call *the core graphs*. For each pair of domains $D_\ell$ and $D_r$ with sizes $d_\ell = |D_\ell|, d_r = |D_r|$, there will be one or more core graphs $G \in \mathcal{G}$. Each graph $G \in \mathcal{G}$ is of the form $G = (V_\ell \cup V_r, E)$ where $V_\ell = \{v_1, \ldots, v_{d_\ell}\}$ and $V_r = \{w_1, \ldots, w_{d_r}\}$ are sets of vertices and $E \subseteq V_\ell \times V_r$. For each graph $G$ we create a set of size $k_G$ of triples of the form $T(G) = \{(G, \pi_{\ell i}, \pi_{ri}) \mid 1 \leq i \leq k_G\}$, where $\pi_{\ell i} : V_\ell \leftrightarrow D_\ell$ and $\pi_{ri} : V_r \leftrightarrow D_r$ are bijections labeling the vertices.

We define $\mathcal{T} = \mathcal{T}(\mathcal{G}) = \cup_{G \in \mathcal{G}} T(G)$. For each $T \in \mathcal{T}$, where $T = (G, \pi_\ell, \pi_r)$, we let $L(T)$ be the set of pairs $\{(\pi_\ell(v), \pi_r(w)) \mid (v, w) \in E(G)\}$. We define $\mathcal{L} = \mathcal{L}(\mathcal{T}) = \{L(T) \mid T \in \mathcal{T}\}$, the set of relations induced by the triples.

Given a bipartite graph $G = (V_\ell \cup V_r, E)$, define $\theta_i$ to be the minimum degree of a vertex in the vertex subset $V_i$ where $i \in \{\ell, r\}$.

We say that $\mathcal{G}$ is *degree bound* if for each graph $\theta_\ell > \frac{d_r}{2}$ and $\theta_r > \frac{d_\ell}{2}$.

**Lemma 3.4.** *If $\mathcal{G}$ is degree bound then for any $T_1, T_2 \in \mathcal{T}$ sharing the domain $D = D_{r1} = D_{\ell 2}$, for any $\alpha \in D_{\ell 1}, \gamma \in D_{r2}$, there exists $\beta \in D$ such that $(\alpha, \beta) \in L_1 = L(T_1), (\beta, \gamma) \in L_2 = L(T_2)$.*

*Proof.* Let $v = \pi_{\ell 1}^{-1}(\alpha)$ be the vertex in $G_1$ labeled by $\alpha$. Since $\deg(v) \geq \theta_{\ell 1}$, the number of pairs in $L_1$ which have $\alpha$ as their first element must be greater than $\frac{d}{2}$. Similarly, the pairs in $L_2$ with $\gamma$ as the second element must also cover more than $1/2$ of $D$. Thus, there must exist $\beta$ such that $(\alpha, \beta) \in L_1, (\beta, \gamma) \in L_2$. $\square$

**Corollary 3.5.** *If $\mathcal{G}$ is degree bound on a common domain $D$ then $\mathcal{L}(\mathcal{T})$ is SC-inducing.*

It is not hard to see that without restricting the bijections Lemma 3.4 provides the best possible result.

**Lemma 3.6.** *If $\mathcal{T}$ is not degree bound and arbitrary bijections $\pi_\ell$ and $\pi_r$ are allowed, $\mathcal{L}(\mathcal{T})$ may not be SC-inducing.*

*Proof.* Consider two relations $L_1$ and $L_2$ on a common domain $D$ with graphs $G_1, G_2$ such that $\theta_{\ell 1} \leq \frac{d}{2}$ and $\theta_{r2} \leq \frac{d}{2}$. Let $\alpha \in D_1$ and $\beta \in D_4$ and note that the neighbors of the vertices labeled by these values will each cover at most $1/2$ of $D$. Since we can choose $\pi_{r1}$ and $\pi_{\ell 2}$ arbitrarily, we can ensure that there is no value for $x_2$ consistent with $x_1 = \alpha, x_3 = \beta$. $\square$

Consider the graph on a domain with $d = 9$ shown in Figure 1. If we allow arbitrary bijections $\pi_\ell, \pi_r$ then $\mathcal{L}$ is not SC-inducing, but it is WC-inducing, which the reader may readily verify. The degree of the vertices in this graph is $\theta = 4$, which is less than that allowed by Lemma 3.4. It remains WC-inducing because the structure of this graph distributes the edges in a uniform way.

To generalize this observation, assume a common domain $D$ and consider three constraints induced by $L_1, L_2, L_3$ built from some core graph set $\mathcal{G}$. If we have minimum degree $\theta$ then for any value on the left of $L_1$ there will be $\theta$ values on the right of $L_1$, and if we allow arbitrary bijections, these may match any subset of size $\theta$ on the left of $L_2$. Similarly, any value on the right of $L_3$ will match an arbitrary set of $\theta$ values on the right of $L_2$. To ensure that $\mathcal{G}$ generates an $\mathcal{L}$ that is WC-inducing, all we need is that any such arbitrary pair of $\theta$-sets induces at least one edge in any $G \in \mathcal{G}$. If any pair of subsets fails to induce an edge on some core graph, then the corresponding constraint set is not weakly 4-consistent.

Letting $N_G(v)$ be the neighbors of vertex $v$ in graph $G$, this argument generalizes to the formal statement

**Lemma 3.7.** *On a common domain $D$, if $\forall G \in \mathcal{G}$ the minimum left and right degree is $\theta_\ell = \theta_r = \theta$, then $\mathcal{G}$ generates a WC-inducing $\mathcal{L}$ using arbitrary bijections $\pi_\ell, \pi_r$ iff $\forall G \in \mathcal{G}$, for $i \in \{\ell, r\}$ and for all $S \subset V_i, |S| = \theta, |\bigcup_{v \in S} N_G(v)| > d - \theta$.*

A graph is *$\theta$-regular* if every vertex is of degree $\theta$. A graph is *connected* if there is a path between every pair of vertices. Notice that in a degree bound $\mathcal{G}$ each $G$ is necessarily connected. The following is a special case where $G$ will generate WC-inducing but not SC-inducing $\mathcal{L}$.

**Corollary 3.8.** *Suppose we have a common domain $D$ with $d = 2\theta$ for some $\theta$ and $\mathcal{G}$ is a set of $\theta$-regular connected bipartite graphs. Let $\mathcal{T}$ be a set of triples generated by random pairs of bijections on graphs in $\mathcal{G}$. Then $\mathcal{L}(\mathcal{T})$ is WC-inducing.*

*Proof.* By Lemma 3.7, we must show that for an arbitrary subset $S \subset V_\ell$ where $|S| = \theta$, the neighbors of $S$ contain more than $\theta$ vertices. Suppose not. Note that we must have $\theta^2$ edges on the vertices in $S$, so if there are $\theta$ other distinct endpoints (there cannot be fewer) each of these has $\theta$ edges. It follows that these are the only edges on the neighbors, but then $G$ is not connected. $\square$

Lemma 3.7 can be used to show that $\theta^2 \geq d$. It also follows easily that $G$ must be connected. These bounds are met at $d = 4, \theta = 2$ using the eight cycle as illustrated in Figure 1. For larger $d$ we mostly leave the problem of optimizing $\theta$ as a research project in combinatorial design, although we will provide some insight in the next section. A few other WC-inducing graphs are shown in Figure 1.

## 3.3 Recursive Construction

By Lemmas 3.4 and 3.6 if we do not introduce additional restrictions on the bijections, the best tightness we can achieve while maintaining strong 3-consistency is $t < \frac{d^2}{2}$. If we only require 4-consistency, then Lemma 3.7 may allow significant improvement, although it is unclear exactly how much in general. We will now show how to increase tightness by reducing the number of edges in the graphs using a recursive construction for larger domains that simultaneously restricts the set of bijections.

Given a domain $D$ we partition it into a set of blocks $\mathbf{\Pi} = \{\Pi_1, \dots, \Pi_k\}$ subject to $k \geq 3$ and $|\Pi_j| \geq 3, \forall j$. These latter restrictions are in place because the only SC-inducing graphs on pairs of size two are the complete bipartite graphs. The blocks $\Pi_j$ may vary in size.

Each variable $x_i$ must have exactly one partition $\mathbf{\Pi}_i$ of its domain which will be used in the construction of all constraints on $x_i$. For problems with a common domain it is simplest to have just one partition of the common domain for all variables, which is the construction we will now assume.

Let $V = \{v_1, \dots, v_d\}$. A bijection $\pi : V \leftrightarrow D$ is said to *respect the partition* $\mathbf{\Pi}$ if there is a bijection $\pi^* : \mathbf{\Pi} \leftrightarrow \mathbf{\Pi}$ on the partition such that for each $j, 1 \leq j \leq k$ and for each $i \in \Pi_j$ we have $\pi(v_i) \in \pi^*(\Pi_j)$. Note that to meet this condition, $\pi^*$ must satisfy the condition $|\Pi_j| = |\pi^*(\Pi_j)|$ for each $j$. Thus, partitions with varying sizes of blocks will further restrict the set of respectful bijections.

We construct a $\mathbf{\Pi}$-*bound graph* $G$ as follows. Let $V_{\mathbf{\Pi}\ell} = \{v_1', \dots, v_k'\}$ and $V_{\mathbf{\Pi}r} = \{w_1', \dots, w_k'\}$ be sets of vertices corresponding to the blocks of the partition. Let $G_{\mathbf{\Pi}} = (V_{\mathbf{\Pi}\ell} \cup V_{\mathbf{\Pi}r}, E_{\mathbf{\Pi}})$ be an SC-inducing graph.

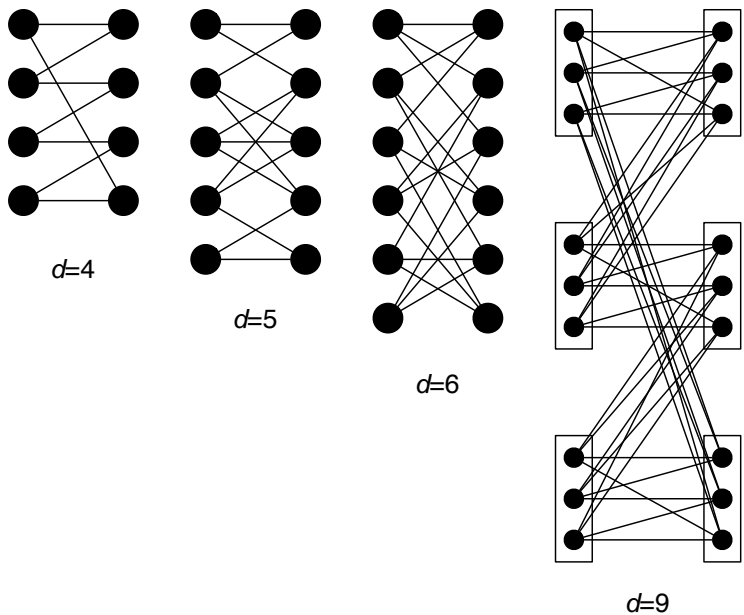

Figure 1: Graphs that are WC-inducing but not SC-inducing for domains of size 4,5,6 and 9. Notice that for $d = 5$ the minimal graph is not regular. The boxes in the graph for $d = 9$ indicate the recursive construction as described in Section 3.3. See also Figure 2.

We assume as in the general construction two sets of vertices $V_\ell = \{v_1, \ldots, v_d\}$ and $V_r = \{w_1, \ldots, w_d\}$. For each edge $(v, w) \in E_{\mathbf{\Pi}}$ we construct a degree bound graph $G_{vw} = (V_\ell[\Pi_v] \cup V_r[\Pi_w], E_{vw})$, where $V_\ell[\Pi_v] = \{v_i : i \in \Pi_v\}$ and $V_r[\Pi_w] = \{w_i : i \in \Pi_w\}$. Recall that degree bound graphs work even when the domains are of differing sizes. We construct the $\mathbf{\Pi}$-bound graph $G = (V_\ell \cup V_r, E)$ by letting $E = \cup_{(v,w) \in E_{\mathbf{\Pi}}} E_{vw}$.

**Lemma 3.9.** *If $\mathcal{G}$ consists of $\mathbf{\Pi}$-bound graphs and all bijections $\pi_\ell, \pi_r$ respect $\mathbf{\Pi}$, then $\mathcal{L}$ is SC-inducing.*

*Proof.* Consider two relations $L_1, L_2$. Since block graphs $G_{\mathbf{\Pi}}$ are SC-inducing, it follows that there is a path between any block on the left graph and any block on the right under the block permutations $\pi^*$ implied by respectful bijections. But now focusing on the edges and bijections restricted to the blocks on such a path, since these induce degree bound graphs, the proof is completed by an argument similar to the proof of Lemma 3.4. □

Figure 1 shows a graph generated by this method for $d = 9$. (For $d = 3$, there is only one minimal degree bound graph up to isomorphism.) Figure 2 shows the construction of two constraints using two copies of this graph. The bijections respect the block structure, and so the constraint is SC-inducing.

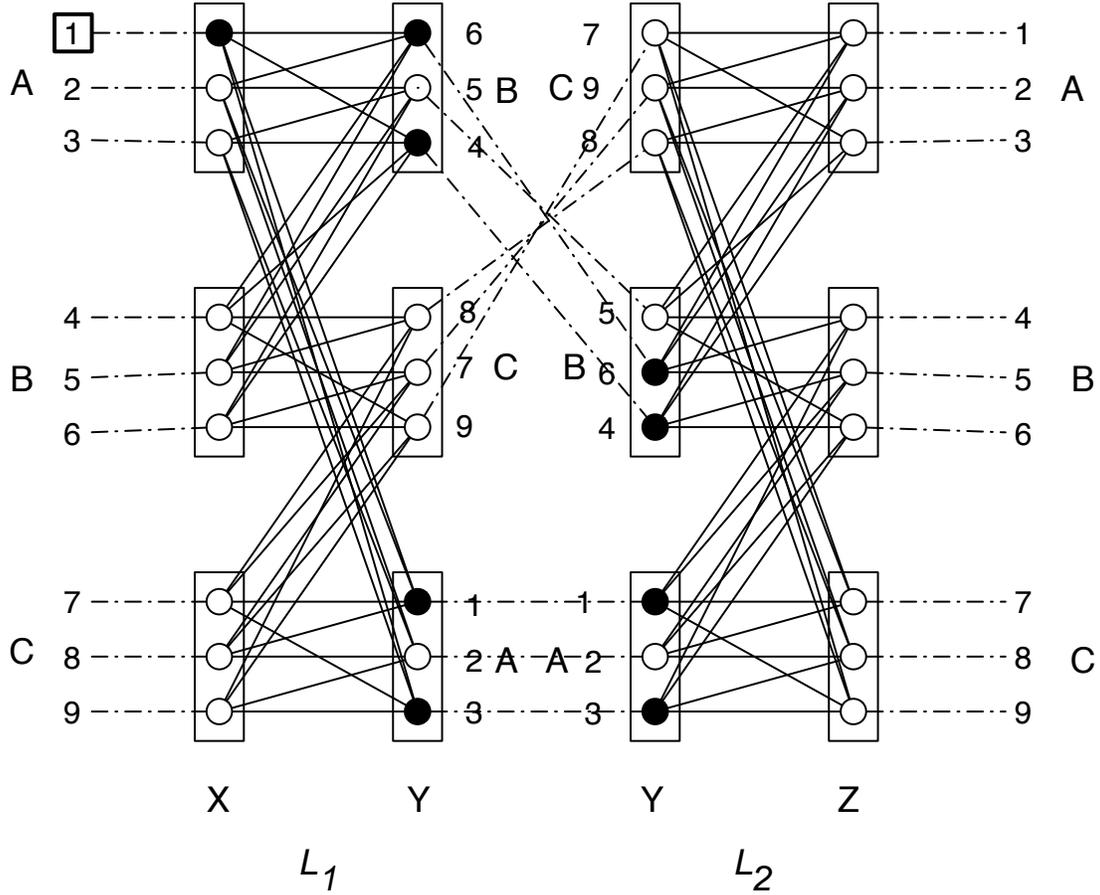

Figure 2: An illustration of a strongly 3-consistent pair constructed using a $3 \times 3$ recursive construction on a domain of size 9. For simplicity, we use identity bijections on the far left and right. In the center, the dashed lines represent the bijection on the shared domain induced by $\pi_{r1}$ and $\pi_{\ell 2}$. $\pi_{r1}$ and $\pi_{\ell 2}$ respect the blocks, but rearrange the values at both the block and vertex levels. We highlight value 1 on the left and the vertices connected to it up to the left side of $L_2$, where there are 2 vertices in each of two blocks. These are sufficient to connect to any value on the right.

For very large domains, we notice that we may have many blocks in $\Pi$. Since $G_{\Pi}$ is SC-inducing, we can recursively use the construction to build $G_{\Pi}$. We call this construction the *recursive $\Pi$-bound construction*. For example, similar to the graph for $d = 9$ in Figure 1, we see that for domains of size $d = 3^i$ this recursive construction lets us use graphs with degree $2^i$. Thus, tightness can be as high as $d^2 - d^{1+\log_3 2}$ on SC-inducing graphs.

We note that it is also possible to recursively partition very large blocks, resulting in possibly non-uniform depth recursive constructions, provided all permutations are block respecting in the appropriate sense. We leave this as an exercise as the notation gets messy, and applications appear limited.

We also claim without proof that if all blocks in $\mathbf{\Pi}$ are of the same size, and we make $G_{\mathbf{\Pi}}$ WC-inducing and replace the degree bound subgraphs with WC-inducing graphs, then provided we use $\mathbf{\Pi}$-respecting bijections, $\mathcal{G}$ is WC-inducing. We call such graphs *weakly* $\mathbf{\Pi}$-*bound*. As before, we can recursively construct $G_{\mathbf{\Pi}}$ to get *recursive weakly* $\mathbf{\Pi}$-*bound* graphs. For $d = 4^i$ using a recursion based on the $d = 4$ case in Figure 1, we can have weak 4-consistency where the core graphs have degree $2^i$; that is, tightness in this case can be as high as $d^2 - d^{3/2}$.

Although there is increased tightness using these recursive constructions, there is a down side in that we greatly reduce the set of allowed bijections. For example, to respect the block structure of the graph on $d = 9$ shown in Figure 2, we have at most $(3!)^4 = 1296$ distinct bijections on either side, compared to $9! = 362880$ unrestricted bijections. We also note that the recursive constructions have limited practical application since it is expensive to express tight constraints in an explicit way for such large domains.

### 3.4 Limits to Tightness and Complexity Tradeoffs

Can we get even tighter constraints with exponential resolution complexity? The discussion so far indicates we are near the limits using our consistency inducing constructions, so we will need something else. The following construction illustrates that it is possible to maintain exponential complexity while having even higher tightness, but it is trivial in the sense that it is really only exponential in a linear sized embedded subproblem.

Let us embed 3-coloring in a domain of size $d > 3$. If we wish to construct a CSP version of $k$-coloring for graphs, we start with a coloring domain $K = \{1, \ldots, k\}$. We create one graph $G = (V_\ell \cup V_r, E) \in \mathcal{G}$, with edge set $E = V_\ell \times V_r \setminus \{(v_i, w_i) \mid 1 \leq i \leq k\}$. We restrict the bijections to $\pi_\ell(v_i) = i, \pi_r(w_i) = i, 1 \leq i \leq k$, which we refer to as the *identity bijections* or *identities*, $\iota_\ell, \iota_r$ for short. So there is only one relation in $\mathcal{L} = \{L(G, \iota_\ell, \iota_r)\}$. To complete the $k$-coloring implementation, we set $t = k$, the maximum possible value for $t$. To embed 3-coloring in a domain of size $d$, let $G$ be constructed as in the 3-coloring construction on the sub-domain $\{1, 2, 3\}$ and then pad $G$ with independent vertices $\{v_i, w_i \mid 4 \leq i \leq d\}$. We construct one triple $(G, \iota_\ell, \iota_r)$ and again we have one relation in $\mathcal{L}$. We set tightness to its maximum possible value, $t = d^2 - 6$.

We conjecture this to be the maximal possible tightness with exponential complexity, but it is of little relevance for larger $d$. For this $\mathcal{L}$, on any instance from $\mathcal{B}_{n,m}^{d,t}[\mathcal{L}]$ any reasonable algorithm will simply reduce it to 3-coloring, and will thus exhibit the exponential complexity of the 3-coloring threshold.

We see that maximizing tightness is certainly not the only consideration, and probably not the most important. We feel it is more important to consider the kinds of structure that one can embed in a model, while avoiding generating trivial instances. Our model expands the framework wherein this can be done over previous models.

## 3.5 Implementation Issues

The experiments conducted by Gao and Culberson (2004) and many of those in this paper use a domain size of $d = 4$ and the 8 cycle WC-inducing graph shown in Figure 1. We should point out that in the experiments conducted by Gao and Culberson (2004) we fixed the bijection $\pi_\ell = \iota_\ell$ and randomly assigned only the right hand set $\pi_r$. Although this is adequate to satisfy the WC-inducing conditions, it does not include all possible $L$, since for example for any $L$ generated that way there will be an $\alpha$ such that both $(1, \alpha)$ and $(2, \alpha)$ are in $L$. If during the construction of constraints the variables are ordered in a consistent way, for example with the smaller index always on the left side of the constraint, then this may introduce an overall asymmetry in the CSP.

In the following, we report a new series of experiments in which we used the randomly paired bijection approach.

The best degree bound graphs, in the sense of maximizing $t$, are those in which all vertices are of the minimum possible degree, that is $\theta$-regular graphs. For our purposes, there is no requirement that these graphs be uniformly selected from all possible such graphs, and it is not too hard to find various techniques to get a variety of such graphs. For graphs conforming to Corollary 3.8 for example, one can first generate a cycle alternating between vertices of $V_\ell, V_r$ and then choose a set of maximum matchings from the remaining unused pairs to reach the required degree. These can be found using any of a variety of techniques, for example by finding a 1-factorization of a regular bipartite graph (Bondy & Murty, 1976, Chapter 5). Should one desire to have a uniform sampling for some reason, there is a significant literature on generating random regular graphs of various types, which might serve as a starting point (Wormald, 1999).

## 4. Experiments

In this section, we discuss a series of experiments we have conducted to study the impact of the structural elements introduced in the generalized flawless model on the typical case hardness of problem instances of practical size. In Section 4.1, we discuss an experiment on a Boolean 3-ary CSP model obtained from the widely-used random distribution of CNF formula. In Section 4.2, we report detailed experimental results on the comparison of three different random CSP models.

### 4.1 Randomly-generated 3-ary Boolean CSPs

The observation from this set of experiments on Boolean CSPs with different constraint tightness partly motivated this study on the interplay between the performance of backtracking algorithms and the structural information in the seemingly "structureless" randomly-generated instances. We include the results here since they provide a nice illustration of the effect of an increase in the constraint tightness (hence an increase in the likelihood of the existence of a forcer in a constraint) on the typical case hardness of random CSP instances.

To obtain an instance distribution for the ternary boolean CSP, we start with a random 3-CNF formula in which each clause is generated by selecting three variables uniformly at random without replacement. A clause in a CNF formula is equivalent to a constraint with constraint tightness $t = 1$ over the same set of variables. E.g., $x \vee y \vee \overline{z}$ corresponds to the restriction $\{(0, 0, 1)\}$. We can therefore obtain random instances of Boolean-valued CSPs with a larger constraint tightness by adding more clauses defined over the same set of variables.

Let $\mathcal{F}(n, m)$ be a random 3-CNF formula with $n$ variables and $m$ clauses selected uniformly at random without replacement. We construct a new random 3-CNF formula $\mathcal{F}(n, m, a)$ as follows:

1. $\mathcal{F}(n, m, a)$ contains all the clauses in $\mathcal{F}(n, m)$;

2. For each clause $C$ in $\mathcal{F}(n, m)$, we generate a random clause on the same set of variables of $C$, and add this new clause to $\mathcal{F}(n, m, a)$ with probability $a$.

In fact, $\mathcal{F}(n, m, a)$ is the random Boolean CSP model with an average constraint tightness $1 + a$ and has been discussed by Gao and Culberson (2003). For $a > 0$, it is easy to see that $\mathcal{F}(n, m, a)$ is always strongly 2-consistent, but is not 3-consistent asymptotically with probability 1.

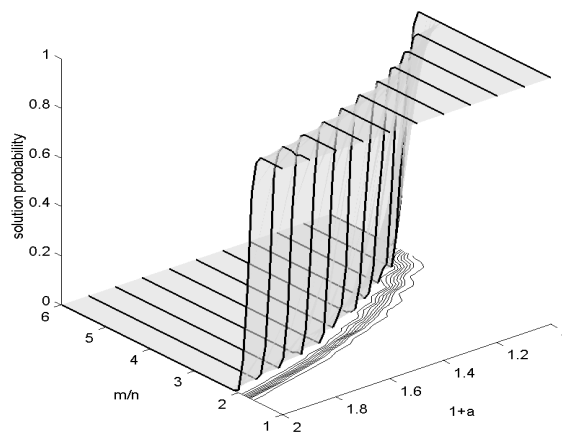

Figure 3: Thresholds for the solution probability in the model $\mathcal{F}(n, m, a)$ with $n = 250$. The z-axis is the solution probability. The axis with the range 1.0 ... 2.0 is for the parameter $1 + a$ and the axis with the range 1.0 ... 6.0 is for the clause density $m/n$.

Figure 4 shows the median of the number of branches used by the SAT solver ZChaff (Zhang et al., 2001) on 100 instances of $\mathcal{F}(n, m, a)$ with $n = 250$. Figure 3 shows the solution probability of the same model. As expected, an increase in the tightness results in a shift of the location of the hardness peak toward smaller $m/n$. More significant is the magnitude of the decrease of the hardness as a result of a small increase in the constraint tightness. The instances hardness varies so dramatically that it is hard to illustrate the difference for all the constraint tightness values from $t = 1.0$ to $t = 2.0$ using either the original scale or the log scale. This explains the scale scheme used in Figure 4.

The reason for such dramatic change can be explained as follows. If we randomly generate constraints with tightness $t > 1$ then for each constraint, there is a positive (fixed) probability that the restriction set will be something such as $\{(1, 0, 0), (0, 0, 0)\}$. This constraint is not 3-consistent, as the partial assignment $y = 0, z = 0$ has no consistent extension to $x$. In effect, this constraint induces the binary constraint on $y, z$ with the restriction $\{(0, 0)\}$. If there are enough such constraints, then the induced binary constraints create an unsatisfiable subproblem, and since the domain is also Boolean, this subproblem can be detected in polynomial time. Upper bounds on $m/n$ have been established for the existence of the easy unsatisfiable subproblems in $\mathcal{F}(n, m, a)$. For example, we know that the upper bounds on $m/n$ for $\mathcal{F}(n, m, a)$ to have an exponential resolution complexity are respectively 23.3 if $a = 0.1$ and 11.7 if $a = 0.2$ (Gao & Culberson, 2003). Since the ratio of constraints to variables $m/n$ considered in the experiment are well below these bounds above which embedded 2SAT subproblems appear with high probability, it seems that the impact of forcers on the instance hardness goes beyond simply producing higly-structured 2-SAT-like embedded easy subproblems. We will see a similar effect in the next subsection on non-Boolean valued CSP instances.

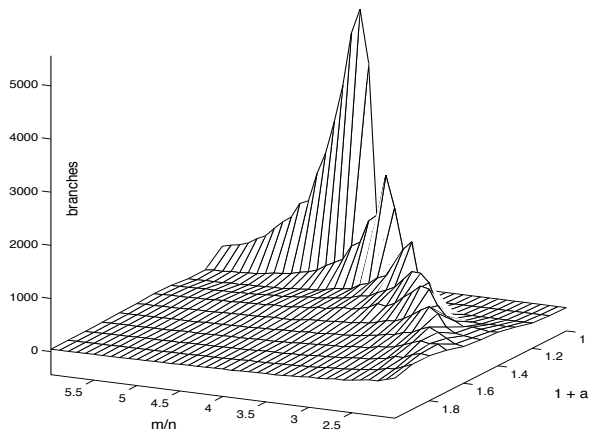

Figure 4: Effects of an increase in the constraint tightness on the instance hardness for $\mathcal{F}(n, m, a)$ with $n = 250$. The z-axis is the median number of branches with the first three highest peaks scaled to $1/40$, $1/10$, and $1/4$ respectively. The axis with the range 1.0 ... 1.8 is for the parameter $1 + a, 0 \leq a \leq 0.8$ and the axis with the range 2.5 ... 6 is for the clause density $m/n$.

## 4.2 Random Binary CSP Models: the Significance of Structures

This set of experiments is designed to investigate whether introducing structural elements that enforce constraint consistency in random CSPs leads to a significant increase in the typical case hardness of instances of practical size. It should be mentioned that the purpose of the experiments is not to compare and rank the relative merits of the solvers we have used in the experiments. Neither is it our intention to exhaust all the available solvers and implementation techniques to solve this set of problem instances.

The three random CSP models we consider are $\mathcal{B}_{n,m}^{d,t}$ (the Model B), $\mathcal{B}_{n,m}^{d,t}[\mathcal{M}]$ (the flawless model), and the generalized flawless model $\mathcal{B}_{n,m}^{d,t}[\mathcal{L}]$ with different domain size and different consistency-inducing core graphs. Randomly-generated instances from these models are encoded as CNF formulas and solved by the SAT solver ZChaff [1]. Also included are some experiments on the comparison of ZChaff, SatZ (Li & Anbulagan, 1997) [2], and a CSP solver based on forward checking (FC) and maintaining arc consistency (MAC) with a static variable order[3].

It looks unnatural that we have primarily tested random CSP instances by converting them to SAT instances and using a SAT solver to solve them. This is justified by the following considerations. First, all of the existing research on the resolution complexity of random CSPs has been carried out by studying the resolution complexity of a SAT encoding of CSPs as described in Section 3. We use the same encoding in the experiments. Secondly, it has been shown that as far as the complexity of solving unsatisfiable CSP instances is concerned, many of the existing CSP algorithms can be efficiently simulated by the resolution system of the corresponding SAT encodings of the CSPs (Mitchell, 2002a).

Experimental comparisons were conducted for the three CSP models with domain size (d = 4, 5, 6, and 9) and different values of constraint tightness. The generalized flawless CSP models are constructed using the WC-inducing core graphs as described in Figure 1. In the next two subsections, we focus on the results for the domain size d = 4 and 9. The experiments were carried out on machines with AMD Athlon (tm) Processor (Model 3700, Frequency 2.4GHz, Main Memory 1GB with 1MB Cache) running Linux. The following setup was used in all of our experiments: the sample size for each parameter point is 100; the cutoff time for all the solvers is 1800 seconds of CPU time.

### 4.2.1 THE CASE OF $d = 4$

For the case of $d = 4$, the $\mathcal{B}_{n,m}^{d,t}[\text{WC}]$ we use in the experiment is based on the WC-inducing core graph shown in Figure 1, a connected 2-regular bipartite graph (or an 8-cycle). Note that for this core graph, the maximum possible constraint tightness is 8.

To observe the detailed behavior of the models, we first fixed the constraint tightness to $t = 6$ and the number of variables to $n = 500$. Figure 5 plots the solution probability as a function of the ratio of constraints to variables for the three CSP models. From the experimental data, we observed that the phase transition of $\mathcal{B}_{n,m}^{d,t}[\text{WC}]$ is much sharper than those of $\mathcal{B}_{n,m}^{d,t}[\mathcal{M}]$ and $\mathcal{B}_{n,m}^{d,t}$[4].

More importantly, instances of $\mathcal{B}_{n,m}^{d,t}[\text{WC}]$ at the phase transition are clearly much harder than those of $\mathcal{B}_{n,m}^{d,t}$ and $\mathcal{B}_{n,m}^{d,t}[\mathcal{M}]$. Figures 6, 7, and 8 show the 50th percentile, 20th percentile, 10th percentile, and 5th percentile of the number of branches and running time in seconds for ZChaff to solve randomly-generated CSP instances from the three models. As can be seen, instances drawn from $\mathcal{B}_{n,m}^{d,t}[\text{WC}]$ are consistently much harder than those from $\mathcal{B}_{n,m}^{d,t}$ and $\mathcal{B}_{n,m}^{d,t}[\mathcal{M}]$ in terms of both the size of the search tree and the running time and for satisfiable instances (Figure 8) as well as unsatisfiable instances.

---

1. Available at http://www.princeton.edu/~chaff/zchaff.html
2. Available at http://www.laria.u-picardie.fr/~cli/EnglishPage.html
3. Available at http://ai.uwaterloo.ca/~vanbeek/software/software.html
4. The reader should be aware that this model doesn't show a phase transition in the limit (as problem size approaches infinity).

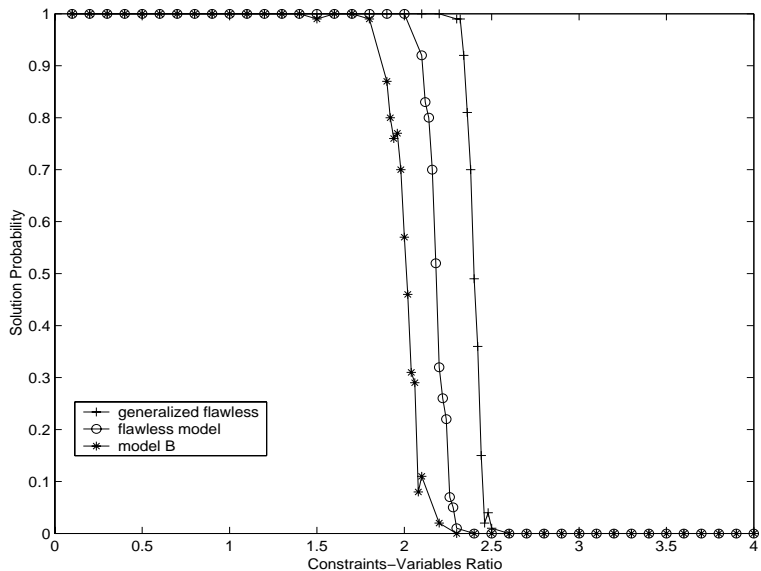

Figure 5: Solution probability as a function of the ratio of constraints to variables for the three random CSP models with $n = 500, t = 6$. For the generalized flawless model, $\mathcal{L}$ is the set of connected 2-regular bipartite graphs. The y-axis is for the solution probability and x-axis for the ratio of constraints to variables $m/n$. Sample size for each data point is 100.

Careful readers may have noticed that for the model $\mathcal{B}_{n,m}^{d,t}[\text{WC}]$, the number of branches has an obvious secondary peak before the phase transition (Figure 6 and Figure 8), an "anomaly" that we have reported in our previous paper (Gao & Culberson, 2004). A detailed look at the experimental data reveals that the secondary peak, though persistent in a statistical sense, appears only for the number of branches; We noticed that the measure of running time does not have such a secondary peak. We have tried several approaches to understand such a secondary peak, and finally conclude that it is a solver-dependent behavior caused the branching heuristics, the CNF encoding, and the level of the local consistency enforced by the CSP models. We provide some further discussion of this phenomenon in the next subsection.

For the case of $d = 4$, we also studied the behavior of the CSP models with different constraint tightness values. The results are reported in Tables 1 and 2 where we show the maximum over all the ratio of constraints to variables of the median of the number of branches for the three CSP models with the constraint tightness ranging from $t = 5$ to $t = 8$. The results are consistent with those observed in the experiments on Boolean CSPs, showing that an increase in constraint tightness has a significant impact on the typical case instance hardness. It is also clear that for all the constraint tightness values, instances from the generalized flawless model $\mathcal{B}_{n,m}^{d,t}[\text{WC}]$ are significantly harder than those from the two classical models. In the case of $t = 5$, instances drawn from $\mathcal{B}_{n,m}^{d,t}[\text{WC}]$ cannot be solved within the time limit of 1800 seconds even before the solubility phase transition.

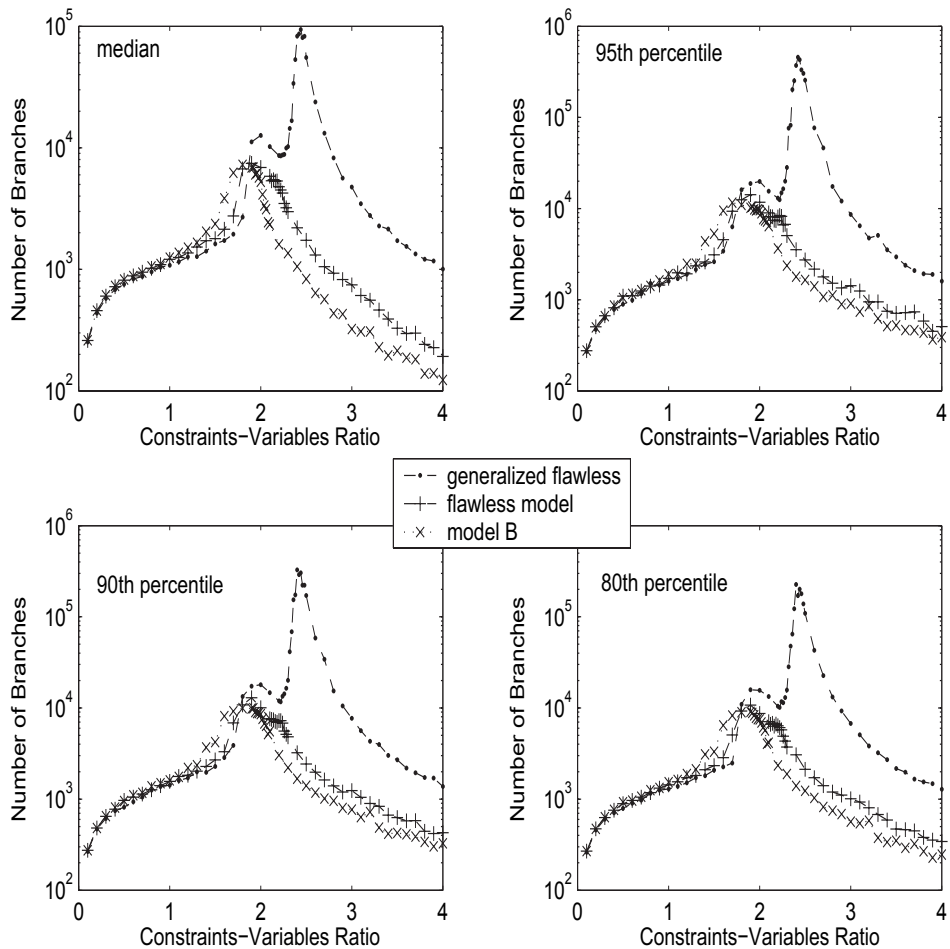

Figure 6: Number of branches used by ZChaff to solve instances from the three random CSP models with $n = 500, t = 6, d = 4$. For the generalized flawless model $\mathcal{B}_{n,m}^{d,t}[\text{WC}]$, the core graph is the connected 2-regular bipartite graph described in Figure 1.

To further confirm that enforcing constraint consistency increases the typical case hardness, we tested two other backtracking solvers in our experiments: SatZ and the CSP solver with FC + MAC and a static variable ordering.

In this set of experiments, the problem size is $n = 300$ and we run all the solvers on the same instances to reduce variance. We summarize the results in Tables 3 and 4.

For problem size n = 300, ZChaff and SatZ have little difficulty in solving randomly-generated instances from all the three models except that in some rare cases, SatZ can get stuck on instances generated with the ratio of constraints to variables well below the threshold. Although the CSP solver works directly on the CSP instances, it does not perform as well as the SAT solvers working on the encoded instances. Given the long history of development of SAT solvers, with significant branch selection heuristics of SatZ and clause-learning (nogood recording) in ZChaff, this should not be surprising. In particular, the design of $\mathcal{B}_{n,m}^{d,t}[\text{WC}]$ can be seen as rendering the consistency checking of the CSP solver less effective. Still, the better performance of the SAT solvers indicates there is still some structure to exploit.

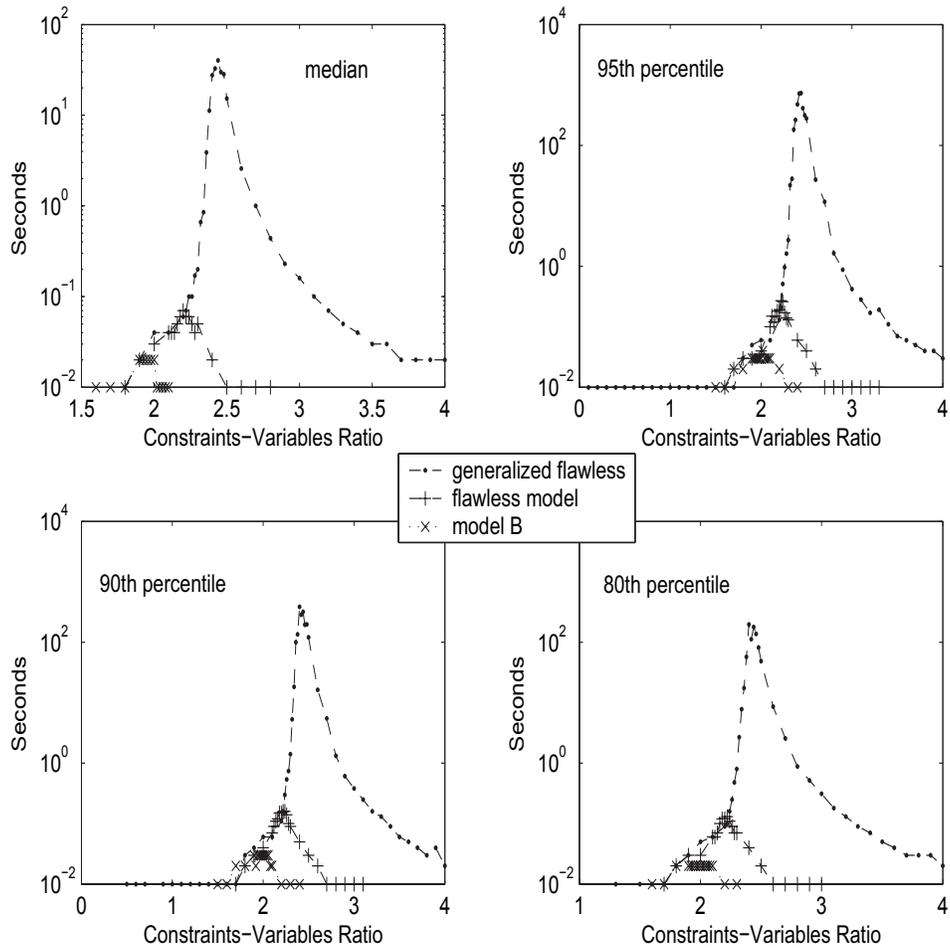

Figure 7: User CPU time in seconds to solve instances from the three random CSP models with $n = 500, t = 6, d = 4$. For the generalized flawless model $\mathcal{B}_{n,m}^{d,t}[\text{WC}]$, the core graph is the connected 2-regular bipartite graph described in Figure 1.

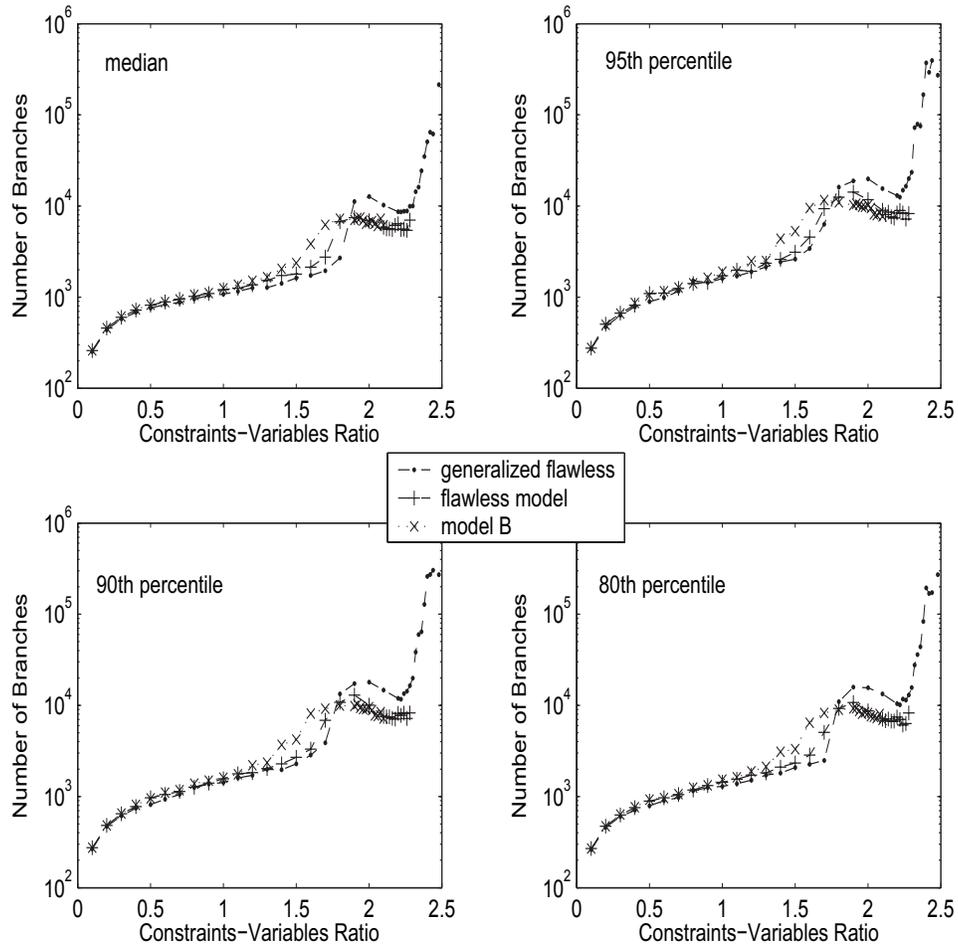

Figure 8: Number of branches used by ZChaff to solve satisfiable instances from the three random CSP models with $n = 500, t = 6, d = 4$. For the generalized flawless model $\mathcal{B}_{n,m}^{d,t}[\text{WC}]$, the core graph is the connected 2-regular bipartite graph described in Figure 1.

| | $\mathcal{B}_{n,m}^{d,t}$ | | $\mathcal{B}_{n,m}^{d,t}[\mathcal{M}]$ | | $\mathcal{B}_{n,m}^{d,t}[\text{WC}]$ | |
|---|---|---|---|---|---|---|
| $(n,t)$ | *branches* | $m/n$ | *branches* | $m/n$ | *branches* | $m/n$ |
| $(500,5)$ | 10673 | 2.30 | 20271 | 2.90 | ? | $> 3.10$ |
| $(500,6)$ | 7243 | 1.80 | 7446 | 1.90 | 94281 | 2.44 |
| $(500,7)$ | 4440 | 1.40 | 5123 | 1.50 | 13813 | 2.00 |
| $(500,8)$ | 1853 | 0.90 | 4113 | 1.20 | 9163 | 1.60 |

Table 1: The maximum, over all $\frac{m}{n}$, of the median of the number of branches of ZChaff on 100 random instances of the three random CSP models , where the domain size $d = 4$, tightness $t = 5, 6, 7, 8$. For the generalized flawless model $\mathcal{B}_{n,m}^{d,t}[\text{WC}]$, the core graph is the connected 2-regular bipartite graph described in Figure 1.

| | $\mathcal{B}_{n,m}^{d,t}$ | | $\mathcal{B}_{n,m}^{d,t}[\mathcal{M}]$ | | $\mathcal{B}_{n,m}^{d,t}[\text{WC}]$ | |
|---|---|---|---|---|---|---|
| $(n,t)$ | *seconds* | $m/n$ | *seconds* | $m/n$ | *seconds* | $m/n$ |
| $(500,5)$ | 0.08 | 2.80 | 1.79 | 2.90 | $> 1800$ | $> 3.10$ |
| $(500,6)$ | 0.02 | 2.00 | 0.02 | 1.90 | 40.3 | 2.44 |
| $(500,7)$ | 0.01 | 1.40 | 0.02 | 1.70 | 1.08 | 2.00 |
| $(500,8)$ | 0.00 | 0.90 | 0.00 | 1.00 | 0.26 | 1.60 |

Table 2: The maximum, over all $\frac{m}{n}$, of the median running time in seconds of ZChaff on random instances of the three random CSP models , where the domain size $d = 4$, tightness $t = 5, 6, 7, 8$. For the generalized flawless model $\mathcal{B}_{n,m}^{d,t}[\text{WC}]$, the core graph is the connected 2-regular bipartite graph described in Figure 1.

### 4.2.2 The Case of $d = 9$

To further study the robustness of our method and to seek explanations on the double peaks, we consider the case of domain size $d = 9$. It turns out that for $d = 9$ we are able to construct a variety of random CSP models that have significantly different behavior. We consider a collection of 5 random CSP models:

1. $\mathcal{B}_{n,m}^{d,t}$ (model B),

2. $\mathcal{B}_{n,m}^{d,t}[\mathcal{M}]$ (the flawless model),

| | ZChaff | | SatZ | | CSP (FC + MAC) | |
|---|---|---|---|---|---|---|
| $(n, t)$ | *seconds* | *m/n* | *seconds* | *m/n* | *seconds* | *m/n* |
| $(300, 5)$ | 4.16 | 3.10 | 2.50 | 3.10 | 90.21 | 3.10 |
| $(300, 6)$ | 0.33 | 2.50 | 0.31 | 2.50 | 12.11 | 2.50 |
| $(300, 7)$ | 0.11 | 2.00 | 0.11 | 2.00 | 3.10 | 2.00 |
| $(300, 8)$ | 0.03 | 1.70 | 0.03 | 1.70 | 0.60 | 1.70 |

Table 3: Maximum, over all $\frac{m}{n}$, of the median number of branches of ZChaff, SatZ, and a CSP solver with FC + MAC on random instances of $\mathcal{B}_{n,m}^{d,t}[\text{WC}]$ , where $n = 300$, $d = 4$, $t = 5, 6, 7, 8$, and the core graph of $\mathcal{B}_{n,m}^{d,t}[\text{WC}]$ is the connected 2-regular bipartite graph described in Figure 1.

| | ZChaff | | SatZ | | CSP (FC + MAC) | |
|---|---|---|---|---|---|---|
| $(n, t)$ | *seconds* | *m/n* | *seconds* | *m/n* | *seconds* | *m/n* |
| $(300, 5)$ | 0.08 | 2.90 | 0.15 | 2.90 | 2.68 | 3.00 |
| $(300, 6)$ | 0.01 | 2.20 | 0.02 | 2.20 | 0.29 | 2.20 |
| $(300, 7)$ | 0.01 | 1.70 | 0.01 | 1.70 | 0.09 | 1.80 |
| $(300, 8)$ | 0.00 | 1.30 | 0.01 | 1.30 | 0.06 | 1.30 |

Table 4: Maximum, overall $\frac{m}{n}$, of the median running time in seconds of ZChaff, SatZ, and a CSP solver with FC + MAC on random instances of $\mathcal{B}_{n,m}^{d,t}[\mathcal{M}]$, where $n = 300$, $d = 4$, and $t = 5, 6, 7, 8$.

3. $\mathcal{B}_{n,m}^{d,t}[\mathcal{L}_1]$ where $\mathcal{L}_1$ is constructed using an 18 cycle (i.e., a connected 2-regular bipartite graph) with arbitrary bijections,

4. $\mathcal{B}_{n,m}^{d,t}[\mathcal{L}_2]$ where $\mathcal{L}_2$ is constructed using the core graph shown in Figure 1 with arbitrary bijections, and

5. $\mathcal{B}_{n,m}^{d,t}[\mathcal{L}_3]$ where $\mathcal{L}_3$ is constructed using the same core graph in Figure 1 with block-respecting bijections.

Recall that the model $\mathcal{B}_{n,m}^{d,t}[\mathcal{L}_2]$ belongs to the class $\mathcal{B}_{n,m}^{d,t}[\text{WC}]$, while $\mathcal{B}_{n,m}^{d,t}[\mathcal{L}_3]$ belongs to the class $\mathcal{B}_{n,m}^{d,t}[\text{SC}]$. For all of the five models, we fix the constraint tightness to $t = 45$, the maximum possible constraint tightness that can be achieved by a generalized flawless model with the WC-inducing core graph.

These experiments show that the typical-case hardness of randomly generated instances increases with the level of consistency enforced. In Figure 9, we plot the median of the number of branches and the median running time in seconds for ZChaff to solve instances from $\mathcal{B}_{n,m}^{d,t}$, $\mathcal{B}_{n,m}^{d,t}[\mathcal{M}]$, and $\mathcal{B}_{n,m}^{d,t}[\mathcal{L}_1]$. In Figure 10, we show the median of the number of branches and median of the running time in seconds for ZChaff to solve instances from $\mathcal{B}_{n,m}^{d,t}[\mathcal{L}_2]$ and $\mathcal{B}_{n,m}^{d,t}[\mathcal{L}_3]$.

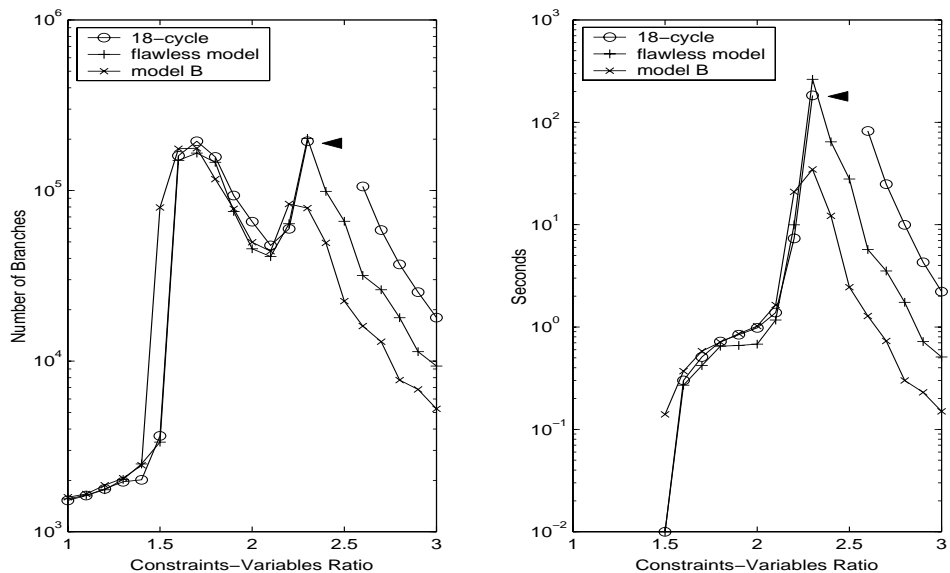

Figure 9: Results on using ZChaff to solve instances from $\mathcal{B}_{n,m}^{d,t}$, $\mathcal{B}_{n,m}^{d,t}[\mathcal{M}]$, and $\mathcal{B}_{n,m}^{d,t}[\mathcal{L}_1]$ with $\mathcal{L}_1$ constructed by an 18 cycle (i.e., a connected 2-regular bipartite graph) and arbitrary bijections. The other parameters used are n = 300, d = 9, and t = 45. For $\mathcal{B}_{n,m}^{d,t}[\mathcal{L}_1]$ with the parameter $m/n = 2.3$, ZChaff failed to solve 45 of the 100 instances. The data (pointed to by solid triangles) for $m/n = 2.3$ are based on the solved instances only.

For $\mathcal{B}_{n,m}^{d,t}$ and $\mathcal{B}_{n,m}^{d,t}[\mathcal{M}]$, ZChaff is able to solve all the instances within the time limit of 1800 seconds for all $m/n$. ZChaff starts to have frequent timeouts when $m/n \geq 2.3$. For $\mathcal{B}_{n,m}^{d,t}[\mathcal{L}_1]$ with $m/n = 2.3$, ZChaff can solve 55 of the 100 instances within 1800 seconds, and 10 of the 55 solved instances are unsatisfiable. At $m/n = 2.4, 2.5$, and $2.6$, ZChaff can solve 42 instances (respectively, 81 instances, 90 instances) all of which are unsatisfiable.

For the models $\mathcal{B}_{n,m}^{d,t}[\mathcal{L}_2]$ and $\mathcal{B}_{n,m}^{d,t}[\mathcal{L}_3]$, we observed that out of 100 instances generated with parameter $m/n = 2.3$, ZChaff can solve 90 instances from $\mathcal{B}_{n,m}^{d,t}[\mathcal{L}_2]$ and 33 instances from $\mathcal{B}_{n,m}^{d,t}[\mathcal{L}_3]$. For $\mathcal{B}_{n,m}^{d,t}[\mathcal{L}_2]$ with $m/n = 2.4$, ZChaff only solved 5 instances. All the solved instances from $\mathcal{B}_{n,m}^{d,t}[\mathcal{L}_2]$ and $\mathcal{B}_{n,m}^{d,t}[\mathcal{L}_3]$ are satisfiable, which is to be expected since this region is well below their threshold. We didn't conduct experiments on $\mathcal{B}_{n,m}^{d,t}[\mathcal{L}_2]$ with $m/n = 2.5$ and $2.6$ and $\mathcal{B}_{n,m}^{d,t}[\mathcal{L}_3]$ with $m/n = 2.4, 2.5$ and $2.6$, but expect that instances from these models will be even harder.

Having assumed that the secondary peak observed in the CSP models with $d = 4$ is algorithm-independent and unique to our models, we had expected that using a larger domain size with a variety of CSP models would be able to help provide a satisfactory explanation of the phenomenon. We designed some experiments to empirically investigate the relations between the appearance of the secondary peak and three characteristics of our

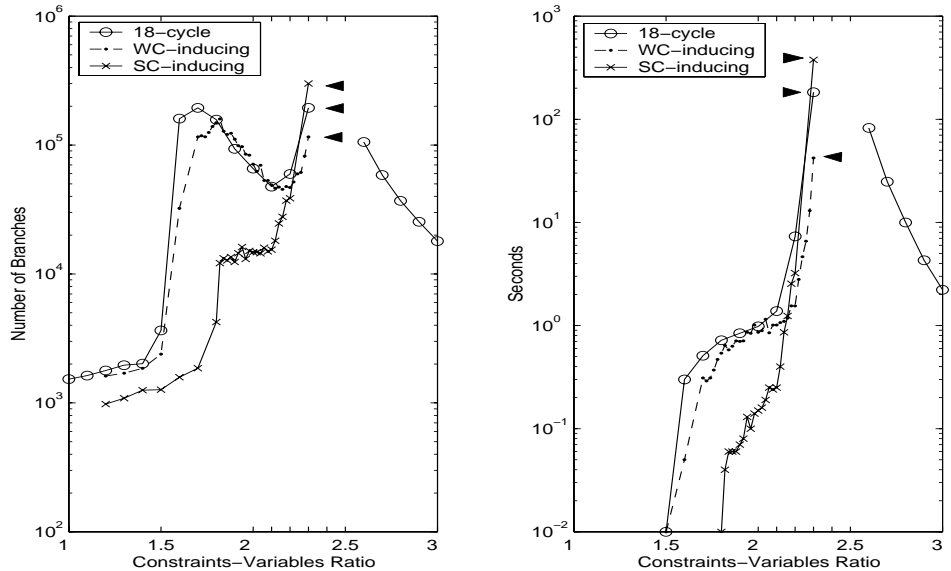

Figure 10: Results on using ZChaff to solve instances from $\mathcal{B}_{n,m}^{d,t}[\mathcal{L}_1]$, weakly 4-consistent model $\mathcal{B}_{n,m}^{d,t}[\mathcal{L}_2]$ and the strongly 3-consistent model $\mathcal{B}_{n,m}^{d,t}[\mathcal{L}_3]$. The other parameters used are d = 9, n = 300, t = 45. ZChaff starts to have frequent timeouts when $m/n \geq 2.3$. Therefore for $m/n = 2.3$, the data (pointed to by solid triangles) are based on the solved instances only. For the model $\mathcal{B}_{n,m}^{d,t}[\mathcal{L}_1]$, 45 of the 55 solved instances are satisfiable and the remaining 10 solved instances are unsatisfiable. For the weakly 4-consistent model $\mathcal{B}_{n,m}^{d,t}[\mathcal{L}_2]$ and the strongly 3-consistent model $\mathcal{B}_{n,m}^{d,t}[\mathcal{L}_3]$, Zchaff solved 90 instances (respectively 33 instances) all of which are satisfiable. For the weakly 4-consistentmodel $\mathcal{B}_{n,m}^{d,t}[\mathcal{L}_2]$ with $m/n = 2.4$, ZChaff can only solve 5 of the 100 instances and these 5 instances are all satisfiable.

CSP models including (1) the number of derangements in the bijections in our constraint construction, (2) the level of the enforced local consistency of the models, and (3) the CNF encoding scheme. Based on a series of results, we conclude that the secondary peak is an algorithm-dependent artifact that is influenced by the selection of the branch heuristics, the level of local consistency enforced by the models, and the CNF encoding scheme. In the following, we briefly summarize these experimental investigations which, we hope, also helps illustrate the flexibility of our framework in constructing an ensemble of problem instances to study the behavior of different algorithms.

1. The Number of Derangements. The number of derangements of a bijection $\pi : V \leftrightarrow D$ is the number of vertices $v_i \in V$ such that $\pi(v_i) \neq i$. If there is no derangement in the bijections in our constraint construction, then the generated instances are always satisfied by the assignments that assign a common domain value to all the variables. Contrary to our initial assumption, we did not observe any significant impact of the number of derangements on the behavior of the solvers except for the extreme cases where we use bijections with no derangement with probability very close to one.

2. The Local Consistency Level. The main difference among the five models with $d = 9$ is the level of the enforced local constraint consistency. $\mathcal{B}_{n,m}^{d,t}$ does not enforce any consistency; $\mathcal{B}_{n,m}^{d,t}[\mathcal{M}]$ enforces arc-consistency; $\mathcal{B}_{n,m}^{d,t}[\mathcal{L}_1]$ guarantees that there is no forcer; $\mathcal{B}_{n,m}^{d,t}[\mathcal{L}_2]$ generates instances that are weakly 4-consistent; and instances from $\mathcal{B}_{n,m}^{d,t}[\mathcal{L}_3]$ are strongly 3-consistent. In addition to the obvious impact on instances hardness, our experiments indicate that the local consistency level in the models contributes to the appearance of the secondary peak of ZChaff. As depicted in Figures 9 and 10, for the problem size $n = 300$, the secondary peak exists in all of the first four models which do not enforce strong 3-consistency, while in the fifth model $\mathcal{B}_{n,m}^{d,t}[\mathcal{L}_3]$ there is no such peak. However, we observed, on some additional trials not reported here, that the secondary peak also exists in $\mathcal{B}_{n,m}^{d,t}[\mathcal{L}_3]$ as we increase the problem size.

3. The CNF Encoding Scheme. In our experiments, the CNF encoding of the CSP instances does not include the clauses that enforce a unique assignment to a CSP variable. While the absence of these clauses does not affect theoretical exponential lower bounds on the resolution complexity, we observed in our experiments mixed impacts of these clauses on the behavior of the solvers including the appearance of the secondary peak of ZChaff. When these clauses are included in the CNF encoding, the number of branches and the running time of SatZ both increase, while for ZChaff the running time has an obvious decrease. It is also interesting that adding these clauses to the CNF encoding also makes the secondary peak of ZChaff even sharper. We believe that this is largely due to the greedy branching heuristics in SatZ and the tradeoff between the search and the inference achieved in ZChaff.

While most of the empirical studies on the phase transitions in the AI literature have focused on the behavior of the solvers around the solubility threshold of the random model, we would like to point out that all of the above observations are made at the ratio of constraints to variables below the respective models' threshold, a parameter region where the selection of the branch heuristics and the no-good recording technique make a big difference.

## 5. Conclusions and Future Directions

Random or semi-random binary constraint instances have been used to test various aspects of CSP solvers by many researchers. To illustrate robustness of algorithms, it is desirable that the instances generated under a model not be trivially unsolvable and a minimal guarantee of interest is that the generators should produce instances that asymptotically have exponential resolution proofs. In this paper we have shown that if we ensure that the constraints are such that the instances are strongly 3-consistent or weakly 4-consistent then this will be the case. In addition we have shown how to create such instances, while allowing for a high constraint tightness, and allowing considerable flexibility in constraint design. In the experimental sections we showed a significant increase in difficulty of instances generated by our model over other models at similar densities and tightness. We also noticed a double peak phenomenon and after further experiments identified it as an artifact of the specific solver, but influenced by the choice of the generation model and the characteristics of the instances. This exemplifies exactly the kind of algorithmic issue that we hope such generators would help researchers identify and explore.

The generation of hard instances has two foci, one in the restriction of the constraints and the other restricting the constraint graph (or hyper-graph). For specific problems, for example independent set (Brockington & Culberson, 1996) or SAT (Bayardo Jr. & Schrag, 1996), it has been observed that techniques such as balancing the degree of the constraint graph to reduce variance increases the difficulty of instances, whether in camouflaging a hidden solution or at the phase transition of semi-random instances. We expect that modifying our generation model by also controlling the graph structure might lead to other harder instances, or other interesting properties that would test the mettle of various algorithms.

For example, suppose we consider the domain $D = \{0, 1, 2, 3\}$ and let a constraint have the set of allowed value pairs $\{(0, 1), (0, 3), (1, 0), (3, 0), (1, 2), (2, 1), (2, 3), (3, 2)\}$. This constraint could not be generated by our system using the eight cycle that enforces weak 4-consistency because this value set induces two four cycles on a pair of domains. Thus, this constraint will not generate weakly 4-consistent CSPs. On the other hand, it is arc consistent and does not contain a forcer, since each vertex in the induced constraint graph has degree two. If we consider a constraint graph with a triangle, then applying this constraint to all three edges will mean there is no satisfying assignment. Since in a random graph with $m/n = c$ there is a positive (bounded) expected number of triangles, there is then a positive expectation that an instance generated allowing this constraint would be trivially unsatisfiable. Thus, it appears that weak 4-consistency is a minimal requirement for exponential complexity on random constraint graphs. Notice that weak 4-consistency not only ensures that every triangle is satisfiable, but it also ensures (by induction basically) that any larger cycle is satisfiable. Thus, speaking in general terms, the only way that a sub-instance can be unsatisfiable is to have the subgraph contain more edges than vertices. But random graph analysis shows this means asymptotically the minimal unsatisfiable sub-instance is of size $O(n)$, and this is of course a key ingredient of the complexity analysis.

Now, suppose we ensure that the constraint graph has girth $g > 3$. Again this technique has been used before on specific problems, for example on graph coloring problems (Culberson, Beacham, & Papp, 1995). We wonder whether or not combining such a graph restriction together with a weak $g + 1$-consistency (weaker than weak 4-consistency) might also produce instances with exponentially long resolution proofs. Note that one difficulty in such an analysis is that we no longer have uniformly random graphs.

Once we start considering the larger picture, involving multiple variables, we naturally must consider $k$-ary constraints. As part of future research we expect to consider extending our model to such cases. Of course some of the experiments on the effects of increasing tightness presented in this paper are on 3-ary constraints (SAT). In fact, our initial foray into tightness started with variations of SAT.

As a final cautionary note, we point out that it is well known that it is possible for CSP (i.e. SAT) instances to have exponential resolution proofs, while being resolvable in polynomial time by other techniques, such as Gaussian elimination. We are not certain our system can produce such instances, but see no explicit reason that prevents it from doing so.

## Acknowledgments

A preliminary version of this paper appeared in the Proceedings of the Tenth International Conference on Principles and Practice of Constraint Programming (CP-2004). We thank the referees for their helpful comments on the conference version and the journal version. Thanks are also given to Dr. K. Xu for his comments on the conference version of the paper. This research has been supported by Natural Sciences and Engineering Research Council Grant No. OGP8053. Yong Gao is supported in part by a UBCO startup grant and an NSERC Discovery Grant RGPIN 327587-06.

## Appendix A: Proofs of the Theorems

In this section, we present more concepts related to the resolution complexity results stated in this paper and prove Theorems 2.1, 3.1, and 3.2.

### A.1 Proof of Theorem 2.1

Before providing the proof for Theorem 2.1, let us first formalize some definitions such as a forcer, a forbidding cycle, and an $r$-flower.

**Definition A.1** (Forcers, Molloy & Salavatipour, 2003). *A constraint $C_f$ with $\mathrm{var}(C_f) = \{x_1, x_2\}$ is called an $(\alpha, \beta)$-forcer if its nogood set is*

$$\{(\alpha, \gamma); \forall \gamma \neq \beta\},$$

*where $\alpha, \beta$, and $\gamma$ are domain values of the involved variables. We say that a constraint $C$ contains an $(\alpha, \beta)$-forcer $C_f$ defined on the same set of variables as $C$ if the nogood set of $C_f$ is a subset of the nogood set of $C$.*

**Definition A.2** (Forbidding cycles and r-flowers, Molloy & Salavatipour, 2003). *An $\alpha$-forbidding cycle for a variable $x_0$ is a set of constraints*

$$C_1(x_0, x_1), C_2(x_1, x_2), \ldots, C_{r-1}(x_{r-2}, x_{r-1}), \text{ and } C_r(x_{r-1}, x_0)$$

544

*such that $C_1(x_0, x_1)$ is an $(\alpha, \alpha_1)$-forcer, $C_r(x_{r-1}, x_0)$ is an $(\alpha_{r-1}, \alpha_r)$-forcer $(\alpha_r \neq \alpha)$, and $C_i(x_{i-1}, x_i)$ is an $(\alpha_{i-1}, \alpha_i)$-forcer $(2 \leq i \leq r-1)$. We call $x_0$ the center variable of the $\alpha$-forbidding cycle.*

*An r-flower $R = \{\mathcal{C}_1, \cdots, \mathcal{C}_d\}$ consists of d (the domain size) forbidding cycles each of which has the length r such that*

1. *$\mathcal{C}_i, 1 \leq i \leq d$, have the same center variable $x$;*

2. *each $\mathcal{C}_i$ is a distinct $\alpha_i$-forbidding cycle of the common center variable $x$; and*

3. *these forbidding cycles do not share any other variables.*

The following facts are straightforward to establish:

1. An r-flower consists of $s = d(r-1) + 1 = dr - d + 1$ variables and $dr$ constraints;

2. The total number of r-flowers is

$$\binom{n}{s} s! (d-1)^d d^{d(r-1)}.$$

3. A constraint in $\mathcal{B}_{n,m}^{d,t}[\mathcal{M}]$ contains an $(\alpha, \beta)$-forcer only if the pair $(\alpha, \beta)$ is one of the tuples that are not considered when selecting the set of nogoods of the constraint.

In the following, we assume that $r = o(\sqrt{n})$. The probability that a constraint contains a forcer and the probability that a random instance of the flawless random CSP model contains an r-flower are given in the following lemma.

**Lemma A.1.** *Consider the flawless random CSP $\mathcal{B}_{n,m}^{d,t}[\mathcal{M}]$ and define $f_e = \frac{\binom{d^2-d-d+1}{t-d+1}}{\binom{d^2-d}{t}}$.*

1. *The probability that a given constraint $C(x_1, x_2)$ contains an $(\alpha, \beta)$-forcer is*

$$\frac{1}{d} f_e. \tag{A.1}$$

2. *Let R be an r-flower and let $c = m/n$,*

$$P\{R \text{ appears in } \mathcal{B}_{n,m}^{d,t}[\mathcal{M}]\} = \Theta(1)(2cf_e)^{dr} \frac{1}{n^{dr}} \frac{1}{d^{dr}}. \tag{A.2}$$

*Proof.* Equation (A.1) follows from the following two observations:

1. $\frac{1}{d}$ is the probability that $(\alpha, \beta)$ is one of the tuples that are not considered when selecting the set of nogoods of a constraint in $\mathcal{B}_{n,m}^{d,t}[\mathcal{M}]$ and

2. $f_e$ is the probability that the $d-1$ tuples, $(\alpha, \gamma), \gamma \neq \beta$, are in the set of $t$ nogoods selected uniformly at random from $d^2 - d$ tuples.

To calculate the probability that a given r-flower $R$ appears in $\mathcal{B}_{n,m}^{d,t}[\mathcal{M}]$, notice that the probability of selecting all the constraint edges in $R$ is

$$
\frac{\binom{N-dr}{cn-dr}}{\binom{N}{cn}} = \frac{cn(cn-1)\cdots(cn-dr+1)}{N(N-1)\cdots(N-dr+1)}
$$

$$
= \Theta(1)\left(\frac{2c}{n}\right)^{dr}
$$

where $N = \binom{n}{2}$. Since for each fixed choice of $dr$ constraint edges in the r-flower, the probability for these constraints to contain the r-flower is $(\frac{1}{d}f_e)^{dr}$, Equation (A.2) follows.

$\square$

**Proof of Theorem 2.1.** Let $c^* = \frac{d}{2f_e}$. We will show that if $c = \frac{m}{n} > c*$, then

$$
\lim_{n\to\infty} P\{\mathcal{B}_{n,m}^{d,t}[\mathcal{M}]\} \text{ contains an r-flower } \} = 1. \tag{A.3}
$$

Let $I_R$ be the indicator function of the event that the $r$-flower $R$ appears in $\mathcal{B}_{n,m}^{d,t}[\mathcal{M}]$ and let

$$
X = \sum_R I_R
$$

where the sum is over all the possible $r$-flowers. Then, $\mathcal{B}_{n,m}^{d,t}[\mathcal{M}]$ contains an r-flower  if and only if $X > 0$.

By Lemma A.1 and the fact that $s = dr - d + 1$, we have

$$
\begin{aligned}
E[X] &= \sum_R E[I_R] \\
&= \Theta(1)\binom{n}{s}s!(d-1)^d d^{d(r-1)}(2cf_e)^{dr}\frac{1}{n^{dr}}\frac{1}{d^{dr}} \\
&= \Theta(1)n(n-1)\cdots(n-s+1)d^{dr}(2cf_e)^{dr}\frac{1}{n^{dr}}\frac{1}{d^{dr}} \\
&= \Theta(1)n^{1-d}(2cf_e)^{dr}.
\end{aligned}
$$

Therefore, if $c > c^*$ and $r = \lambda\log n$ with $\lambda$ sufficiently large, we have $\lim_{n\to\infty} E[X] = \infty$.

If we can show that $E[X^2] \leq E^2[X](1 + o(1))$, then an application of the Chebyshev inequality will establish that $\lim_{n\to\infty} P\{X = 0\} = 0$. To get an upper bound on $E[X^2]$, we need a counting argument to upper bound the number of r-flowers sharing a given number of edges. This is done by considering how the shared edges form connected components (Franco & Gelder, 2003; Gao & Culberson, 2003; Molloy & Salavatipour, 2003). Here, we follow the method used by Molloy and Salavatipour (2003), from which we have

$$
\begin{aligned}
E[X^2] &= \sum_A E[I_A^2] + \sum_A \sum_{B:B\cap A=\oslash} E[I_A I_B] + \sum_A I_A \left( \sum_{i=1}^{s} \sum_{j=1}^{i} N_{ij}(P_{ij})^{dr-i} \right) \\
&= \sum_A E[I_A^2] + \sum_A \sum_{B:B\cap A=\oslash} E[I_A]E[I_B] + \sum_A I_A \left( \sum_{i=1}^{s} \sum_{j=1}^{i} N_{ij}(P_{ij})^{dr-i} \right) \\
&\leq E^2[X] + \sum_A I_A \left( \sum_{i=1}^{s} \sum_{j=1}^{i} N_{ij}(P_{ij})^{dr-i} \right) \quad\quad\quad (A.4)
\end{aligned}
$$

where (1) $N_{ij}$ is the number of the r-flowers that share exactly $i$ constraint edges with $A$ and these $i$ constraints forms $j$ connected components in the constraint graph of $A$; and (2) $(P_{ij})^{dr-i}$ is the probability conditional on $I_A$, that the random CSP contains the $dr-i$ constraints of a specific r-flower. In the work of Molloy and Salavatipour (2003), $N_{ij}$ is upper bounded by

$$
\left( (2+r^2)^d (dr^2)^{j-1} \right)^2 j! n^{s-i-j} d^{s-i-j+d-1},
$$

where $((2+r^2)^d(dr^2)^{j-1})^2 j!$ upper bounds the number of ways to choose and arrange the $j$ shared connected components for two r-flowers; $n^{s-i-j}$ upper bounds the number of ways of choosing the remaining non-shared variables since the number of variables in each of the j shared connected components is at least one plus the number of edges in that shared component; and $d^{s-i-j+d-1}$ upper bounds the number of ways of choosing the forcing values in these non-sharing variables. The shared variables have to take the same forcing values as those in $A$ due to the assumption that $t < d$ (Molloy & Salavatipour, 2003).

Since in our case $d-1 \leq t \leq d^2-d$, it is possible for the shared variables to take different forcing values in different r-flowers. Thus, an upper bound for $N_{ij}$ is

$$
\left( (2+r^2)^d (dr^2)^{j-1} \right)^2 j! n^{s-i-j} d^s.
$$

But in our case, the probability corresponding to $(P_{ij})^{dr-i}$ is

$$
\begin{aligned}
&\frac{\binom{N-dr-(dr-i)}{cn-i-(dr-i)}}{\binom{N-dr}{cn-i}} (\frac{1}{d} f_e)^{dr-i} = \Theta(1) (\frac{cn-i}{N-dr})^{dr-i} (\frac{1}{d} f_e)^{dr-i} \\
&= \Theta(1) (2cf_e)^{dr-i} \frac{1}{n^{dr-i}} \frac{1}{d^{dr-i}}.
\end{aligned}
$$

Therefore, with $c^* = \frac{d}{2f_e}$, we have

$$
\begin{aligned}
&\sum_{i=1}^{s} \sum_{j=1}^{i} N_{ij} (2cf_e)^{dr-i} \frac{1}{n^{dr-i}} \frac{1}{d^{dr-i}} \\
&\leq \sum_{i=1}^{s} \left[ (2+r^2)^{2d} r^{-4} n^{s-i} d^s (2cf_e)^{dr-i} \frac{1}{n^{dr-i}} \frac{1}{d^{dr-i}} \sum_{j=1}^{i} (\frac{d^2 r^4 j}{n})^j \right] \\
&\leq \sum_{i=1}^{s} O(r^{4d-4}) n^{1-d} (2cf_e)^{dr} \frac{(2cf_e)^{-i}}{d^{-i}} O(\frac{r^4}{n}) \\
&\leq E[X] O(r^{4d-4}) O(\frac{r^4}{n}) \sum_{i=1}^{s} (\frac{d}{2cf_e})^i \\
&\leq E[X] O(\frac{r^{4d}}{n}),
\end{aligned}
$$

where the last inequality is because $c > \frac{d}{2f_e}$. From this and formula (A.4), the proof is completed. $\qquad \square$

**Remark A.1.** *The relatively loose upper bound $c^* = \frac{d}{2f_e}$ in the above proof may be improved by a factor of $d$ by making a further distinction among the $r$-flowers that share forcing values at a different number of shared variables. But for the purpose of showing that the flawless random CSP also has potential embedded easy sub-problems, our upper bound for the constraint-variable ratio $c$ is sufficient since the domain size $d$ is a constant.*

## A.2 Proof of Theorems 3.1 and 3.2

Given a CNF formula $\mathcal{F}$, we use $\text{Res}(\mathcal{F})$ to denote the resolution complexity of $\mathcal{F}$, i.e., the length of the shortest resolution refutation of $\mathcal{F}$. The width of deriving a clause $A$ from $\mathcal{F}$, denoted by $w(\mathcal{F} \vdash A)$, is defined to be the minimum over all the resolution refutations of the maximum clause size in the resolution refutation. The width $w(\mathcal{F})$ of a formula $\mathcal{F}$ is the size of the largest clause in it. Ben-Sasson and Wigderson (2001) established a relationship between $\text{Res}(\mathcal{F})$ and $w(\mathcal{F} \vdash \emptyset)$:

$$
\text{Res}(\mathcal{F}) = e^{\Omega(\frac{(w(\mathcal{F} \vdash \emptyset) - w(\mathcal{F}))^2}{n})}.
$$

This relationship indicates that to give an exponential lower bound on the resolution complexity, it is sufficient to show that every resolution refutation of $\mathcal{F}$ contains a clause whose size is linear in $n$, the number of variables.

Let $\mathcal{I}$ be an instance of the CSP and let $\text{CNF}(\mathcal{I})$ be the CNF encoding of $\mathcal{I}$. Mitchell (2002b) provided a framework within which one can investigate the resolution complexity of $\mathcal{I}$, i.e., the resolution complexity of the CNF formula $\text{CNF}(\mathcal{I})$ that encodes $\mathcal{I}$, by working directly on the structural properties of $\mathcal{I}$. Denote by $\text{var}(\mathcal{I})$ the set of CSP variables in $\mathcal{I}$ and $\text{var}(\text{CNF}(\mathcal{I}))$ the set of encoding Boolean variables in $\text{CNF}(\mathcal{I})$. A sub-instance $\mathcal{J}$ of $\mathcal{I}$ is a CSP instance such that $\text{var}(\mathcal{J}) \subset \text{var}(\mathcal{I})$ and $\mathcal{J}$ contains all the constraints of $\mathcal{I}$ whose scope variables are in $\text{var}(\mathcal{J})$. The following crucial concepts make it possible to work directly on the structural properties of the CSP instance when investigating the resolution complexity of the encoding CNF formula.

**Definition A.3** (Implies, Mitchell, 2002b). *Let $C$ be a clause over the encoding Boolean variables in $\mathrm{var}(\mathrm{CNF}(\mathcal{I}))$. We say that a sub-instance $\mathcal{J}$ of $\mathcal{I}$ implies $C$, denoted as $\mathcal{J} \models C$, if and only if for each assignment to the CSP variables satisfying $\mathcal{J}$, the corresponding assignment to the encoding Boolean variables satisfies $C$.*

**Definition A.4** (Clause Complexity, Mitchell, 2002b). *Let $\mathcal{I}$ be a CSP instance. For each clause $C$ defined over the Boolean variables in $\mathrm{var}(\mathrm{CNF}(\mathcal{I}))$, define*

$$\mu(C, \mathcal{I}) = \min\{|\mathrm{var}(\mathcal{J})|; \mathcal{J} \text{ is a sub-instance and implies } C\}.$$

The following two concepts slightly generalize those used by Mitchell (2002b) and by Molloy and Salavatipour (2003) and enable us to have a uniform treatment when establishing resolution complexity lower bounds.

**Definition A.5** (Boundary). *The boundary $\mathcal{B}(\mathcal{J})$ of a sub-instance $\mathcal{J}$ is defined to be the set of CSP variables such that a variable $x$ is in $\mathcal{B}(\mathcal{J})$ if and only if the following is true: If $\mathcal{J}$ minimally implies a clause $C$ defined on some Boolean variables in $\mathrm{var}(\mathrm{CNF}(\mathcal{I}))$, then $C$ contains at least one of the Boolean variables, $x : \alpha, \alpha \in D$, that encode the CSP variable $x$.*

**Definition A.6** (Sub-critical Expansion, Mitchell, 2002b). *Let $\mathcal{I}$ be a CSP instance. The sub-critical expansion of $\mathcal{I}$ is defined as*

$$e(\mathcal{I}) = \max_{0 \leq s \leq \mu(\emptyset, \mathcal{I})} \min_{s/2 \leq |\mathrm{var}(\mathcal{J})| \leq s} |\mathcal{B}(\mathcal{J})| \tag{A.5}$$

*where the minimum is taken over all the sub-instances of $\mathcal{I}$ such that $s/2 \leq |\mathrm{var}(\mathcal{J})| \leq s$.*

The following theorem relates the resolution complexity of the CNF encoding of a CSP instance to the sub-critical expansion of the CSP instance.

**Theorem A.2** (Mitchell, 2002b). *For any CSP instance $\mathcal{I}$, we have*

$$w(\mathrm{CNF}(\mathcal{I}) \vdash \emptyset) \geq e(\mathcal{I}) \tag{A.6}$$

*Proof.* For any resolution refutation $\pi$ of $\mathrm{CNF}(\mathcal{I})$ and $s \leq \mu(\emptyset, \mathcal{I})$, Lemma 1 of Mitchell (2002b) shows that $\pi$ must contain a clause $C$ with

$$s/2 \leq \mu(C, \mathcal{I}) \leq s.$$

Let $\mathcal{J}$ be a sub-instance such that $|\mathrm{var}(\mathcal{J})| = \mu(C, \mathcal{I})$ and $\mathcal{J}$ implies $C$. Since $\mathcal{J}$ minimally implies $C$, according to the definition of the boundary, $w(C) \geq |\mathcal{B}(\mathcal{J})|$. Formula (A.6) follows. □

To establish an asymptotically exponential lower bound on $\mathrm{Res}(\mathcal{C})$ of a random CSP $\mathcal{C}$, it is enough to show that there is a constant $\beta^* > 0$ such that

$$\lim_{n \to \infty} P\{e(\mathcal{C}) \geq \beta^* n\} = 1. \tag{A.7}$$

For any $\alpha > 0$, let $\mathcal{A}_m(\alpha)$ be the event $\{\mu(\emptyset, \mathcal{C}) > \alpha n\}$ and $\mathcal{A}_s(\alpha, \beta^*)$ be the event

$$\left\{ \min_{\frac{\alpha n}{2} \leq |\operatorname{var}(\mathcal{J})| \leq \alpha n} \mathcal{B}(\mathcal{J}) \geq \beta^* n \right\}.$$

Notice that

$$\begin{aligned}
P\{e(\mathcal{C}) \geq \beta^* n\} & \geq P\{\mathcal{A}_m(\alpha) \cap \mathcal{A}_s(\alpha, \beta^*)\} \\
& \geq 1 - P\{\overline{\mathcal{A}_m(\alpha)}\} - P\{\overline{\mathcal{A}_s(\alpha, \beta^*)}\}.
\end{aligned} \tag{A.8}$$

We only need to find appropriate $\alpha^*$ and $\beta^*$ such that

$$\lim_{n \to \infty} P\{\overline{\mathcal{A}_m(\alpha^*)}\} = 0 \tag{A.9}$$

and

$$\lim_{n \to \infty} P\{\overline{\mathcal{A}_s(\alpha^*, \beta^*)}\} = 0. \tag{A.10}$$

Event $\mathcal{A}_m(\alpha^*)$ is about the size of minimally unsatisfiable sub-instances. For the event $\mathcal{A}_s(\alpha^*, \beta^*)$, a common practice is to identify a special subset of the boundaries and show that this subset is large. For different random CSP models and under different assumptions on the model parameters, there are different ways to achieve this. Following Beame et al. (2005), we say a graph $G$ is $(r, q)$-dense if there is a subset of $r$ vertices that induces at least $q$ edges of $G$.

**Proof of Theorem 3.1.** Recall that the constraint graph of $\mathcal{B}_{n,m}^{d,t}[SC]$ is the standard random graph $G(n, m)$.

Since each instance of $\mathcal{B}_{n,m}^{d,t}[SC]$ is strongly $k$-consistent, variables in a minimal unsatisfiable sub-instance $\mathcal{J}$ with $|\operatorname{var}(\mathcal{J})| = r$ must have a vertex degree greater than or equal to $k$, and consequently, the constraint sub-graph $H(\mathcal{J})$ must contain at least $\frac{rk}{2}$ edges. Thus,

$$\begin{aligned}
P\{\overline{\mathcal{A}_m(\alpha^*)}\} & = P\{\mu(\emptyset, \mathcal{B}_{n,m}^{d,t}[SC]) \leq \alpha^* n\} \\
& \leq P\left\{ \bigcup_{r=k+1}^{\alpha^* n} \{G(n, m) \text{ is } (r, rk/2)\text{-dense }\} \right\}.
\end{aligned}$$

Let $\mathcal{B}^k(\mathcal{J})$ be the set of the variables in $\operatorname{var}(\mathcal{J})$ whose vertex degrees are less than $k$. Again, since instances of $\mathcal{B}_{n,m}^{d,t}[SC]$ are always strongly k-consistent, we have $\mathcal{B}^k(\mathcal{J}) \subset \mathcal{B}(\mathcal{J})$ and thus, $|\mathcal{B}(\mathcal{J})| \geq |\mathcal{B}^k(\mathcal{J})|$. Therefore, the probability $P\{\overline{\mathcal{A}_s(\alpha^*, \beta^*)}\}$ can be bounded by

$$P\{\overline{\mathcal{A}_s(\alpha^*, \beta^*)}\} \leq P\{\overline{\mathcal{A}_s^k(\alpha^*, \beta^*)}\}$$

where $\mathcal{A}_s^k(\alpha^*, \beta^*)$ is the event

$$\left\{ \min_{\alpha^* n/2 \leq |\operatorname{var}(\mathcal{J})| \leq \alpha^* n} \mathcal{B}^k(\mathcal{J}) \geq \beta^* n \right\}.$$

Random graph arguments (see, e.g., Beame et al., 2005) show that there exist constants $\alpha^*$ and $\beta^*$ such that $P\{\overline{\mathcal{A}_m(\alpha^*)}\}$ and $P\{\overline{\mathcal{A}_s^k(\alpha^*, \beta^*)}\}$ both tend to 0. Indeed, let $\beta^*$ be such that $\frac{(1-\beta^*)k}{2} > 1$, $c = \frac{m}{n}$, and $N = \frac{n(n-1)}{2}$. We have

$$P\{\overline{\mathcal{A}_m(\alpha^*)}\} \leq P\left\{\bigcup_{r=k+1}^{\alpha^* n} \{G(n,m) \text{ is } (r, rk/2)\text{-dense }\}\right\}$$

$$\leq \sum_{r=k+1}^{\alpha^* n} P\{G(n,m) \text{ is } (r, \frac{rk}{2})\text{-dense}\}$$

$$\leq \sum_{r=k+1}^{\alpha^* n} \binom{n}{r}\binom{\frac{r(r-1)}{2}}{\frac{rk}{2}}\binom{N-\frac{rk}{2}}{m-\frac{rk}{2}}\binom{N}{m}^{-1}$$

$$\leq \sum_{r=k+1}^{\alpha^* n} (\frac{en}{r})^r (\frac{e(r-1)}{k})^{\frac{rk}{2}}(\frac{2c}{n})^{\frac{rk}{2}}$$

$$= \sum_{r=k+1}^{\alpha^* n} \left[\frac{en}{r}(\frac{2ec(r-1)}{kn})^{\frac{k}{2}}\right]^r$$

$$= \sum_{r=k+1}^{\alpha^* n} \left[(\frac{k}{2})^{\frac{k}{2}} e^{\frac{k+2}{2}} c^{\frac{k}{2}} (\frac{r}{n})^{\frac{k-2}{2}}\right]^r$$

$$\leq \sum_{r=k+1}^{\lfloor \log n \rfloor} \left[(\frac{k}{2})^{\frac{k}{2}} e^{\frac{k+2}{2}} c^{\frac{k}{2}} (\frac{\log n}{n})^{\frac{k-2}{2}}\right]$$

$$+ \sum_{r=\lfloor \log n \rfloor}^{\alpha^* n} \left[(\frac{k}{2})^{\frac{k}{2}} e^{\frac{k+2}{2}} c^{\frac{k}{2}} (\alpha^*)^{\frac{k-2}{2}}\right]^{\log n} \tag{A.11}$$

Similarly, we have for $\overline{\beta} = \frac{2\beta^*}{\alpha^*}$,

$$P\{\overline{\mathcal{A}_s^k(\alpha^*, \beta^*)}\} = P\left\{\bigcup_{r=\frac{\alpha^* n}{2}}^{\alpha^* n} \{\exists \text{ a size-r sub-instance } \mathcal{J} \text{ s.t. } |\mathcal{B}^k(\mathcal{J})| \leq \beta^* n\}\right\}$$

$$\leq P\left\{\bigcup_{r=\frac{\alpha^* n}{2}}^{\alpha^* n} \{G(n,m) \text{ is } (r, \frac{r(1-\overline{\beta})k}{2})\text{-dense}\}\right\}$$

$$\leq \sum_{r=\frac{\alpha^* n}{2}}^{\alpha^* n} \left[\left(\frac{2c}{(1-\overline{\beta})k}\right)^{\frac{(1-\overline{\beta})k}{2}} e^{\frac{(1-\overline{\beta})k+2}{2}} (\alpha^*)^{\frac{(1-\overline{\beta})k-2}{2}}\right]^r \tag{A.12}$$

where the second inequality is because of the fact that for a sub-instance $\mathcal{J}$ with size r and $|\mathcal{B}^k(\mathcal{J})| \leq \beta^* n$, its constraint graph contains at least $r - \beta^* n = r - \frac{\alpha^*}{2}\overline{\beta} n \geq r - \overline{\beta} r$ vertices whose degree is at least $k$.

There exist $\alpha^*$ and $\beta^*$ be such that (1) $\frac{2\beta^*}{\alpha^*} < 1$; (2) $\frac{(1-\beta^*)k}{2} > 1$; and (3) the right hand side of formula (A.11) and the right hand side of formula (A.12) both tend to zero. This completes the proof of Theorem 3.1. □

We now prove Theorem 3.2. First from the definition of $\mathcal{B}_{n,m}^{d,t}[WC]$, we have the following

**Lemma A.3.** *For the random CSP $\mathcal{B}_{n,m}^{d,t}[WC]$, we have*

1. *Every sub-instance whose constraint graph is a cycle is satisfiable;*

2. *for any path of length $\geq 3$, any compatible assignments to the two variables at the ends of the path can be extended to assignments that satisfy the whole path.*

In an effort to establish exponential lower bounds on the resolution complexity for a classical random CSP models with a tightness higher than those established by Mitchell (2002b), Molloy and Salavatipour (2003) introduced a collection of sub-instances, denoted here as $\mathcal{B}_M(\mathcal{J})$, and used its size to give a lower bound on the size of the boundary. For binary CSPs whose constraints are arc-consistent and contain no forcer, $\mathcal{B}_M(\mathcal{J})$ consists of two parts: $\mathcal{B}_M^1(\mathcal{J})$ and $\mathcal{B}_M^2(\mathcal{J})$, defined respectively as follows:

1. $\mathcal{B}_M^1(\mathcal{J})$ contains the set of single-edge sub-instances $\mathcal{X}$, i.e., $|\text{var}(\mathcal{X})| = 2$, such that at least one of the variables has a vertex degree 1 in the original constraint graph;

2. $\mathcal{B}_M^2(\mathcal{J})$ contains the set of sub-instances $\mathcal{X}$ whose induced constraint graph is a pendant path of length 4, i.e., a path of length 4 such that no vertex other than the endpoints has a vertex degree greater than 2 in the original constraint graph.

It can be shown that

**Lemma A.4.** *For any weakly path-consistent CSP sub-instance $\mathcal{J}$, we have*

$$|\mathcal{B}(\mathcal{J})| \geq |\mathcal{B}_M^1(\mathcal{J})| + \frac{|\mathcal{B}_M^2(\mathcal{J})|}{4}.$$

*Proof.* The variable with degree one in any sub-instance in $\mathcal{B}_M^1(\mathcal{J})$ has to be in $\mathcal{B}(\mathcal{J})$; At least one internal variable in any pendant path $\mathcal{B}_M^2(\mathcal{J})$ has to be in $\mathcal{B}(\mathcal{J})$. It is possible that several pendant paths of length 4 share a common internal variable that is in $\mathcal{B}(\mathcal{J})$, e.g., in a very long pendant path. But a variable can only appear in at most three pendant paths of length 4. □

With the above preparations, the proof provided for Theorem 1 of Molloy and Salavatipour (2003) readily applies to our case. To make this report self-contained, we give the proof below.

**Proof of Theorem 3.2.** By Lemma A.3, any minimally unsatisfiable sub-instance $\mathcal{J}$ is such that (1) its constraint graph cannot be a single cycle; and (2) $\mathcal{B}_M(\mathcal{J})$ is empty since $|\mathcal{B}_M^1(\mathcal{J})| = 0$ and $|\mathcal{B}_M^2(\mathcal{J})| = 0$ for a minimally unsatisfiable sub-instance. According to Lemma 11 of Molloy and Salavatipour, the constraint graph of $\mathcal{J}$ has at least $(1 + \frac{1}{12})\text{var}(\mathcal{J})$ edges. Therefore, due to the locally sparse property of random graphs, there is a constant $\alpha^* > 0$ such that formula (A.9) holds, i.e.,

$$\lim_{n \to \infty} P\{\overline{\mathcal{A}_m(\alpha^*)}\} = 0.$$

To establish formula (A.10), due to Lemma A.4 we have

$$P\{\mathcal{A}_s(\alpha^*, \beta^*)\} \geq P\{\mathcal{A}_{s,M}(\alpha^*, \beta^*)\}$$

where $\mathcal{A}_{s,M}(\alpha^*, \beta^*)$ is the event

$$\left\{ \min_{\alpha^* n/2 \leq |\mathrm{var}(\mathcal{J})| \leq \alpha^* n} |\mathcal{B}_M(\mathcal{J})| \geq \beta^* n \right\}.$$

Now suppose on the contrary that for any $\zeta > 0$, there is a sub-instance $\mathcal{J}$ with $\alpha^* n/2 \leq |\mathrm{var}(\mathcal{J})| \leq \alpha^*$ such that $|\mathcal{B}_M^1(\mathcal{J})| + |\mathcal{B}_M^2(\mathcal{J})| \leq \zeta n$. Then, from Lemmas 10 and 11 of Molloy and Salavatipour, the constraint graph of $\mathcal{J}$ contains only cycle components, where Lemma 11 of Molloy and Salavatipour asserts that the edges-to-vertices ratio of the constraint graph of $\mathcal{J}$ has to be bigger than one. If we remove all the cycle components from the constraint graph of $\mathcal{J}$, the edges-to-vertices ratio of the remaining graph becomes even bigger. But this is impossible because the constraint graph of $\mathcal{J}$, and hence the remaining graph, has less than $\alpha^* n$ vertices.

It is well-known that w.h.p. a random graph has fewer than $\log n$ cycle components of length at most 4; for the random graph $G(m, n)$ with $m/n = c$ being constant, the number of cycle components with a fixed length has asymptotically a Poisson distribution (Bollobas, 2001). Thus, the number of variables that are in cycle components of length 4 is at most $4 \log n$. Since any cycle component of length $l > 4$ contains $l$ pendant paths of length 4, the total number of variables in cycle components of length greater than 4 is at most $|\mathcal{B}_M^2(\mathcal{J})| < \zeta n$. Therefore, we have $\mathrm{var}(\mathcal{J}) < \zeta n + 4 \log n < \alpha^* n/2 \leq \mathrm{var}(\mathcal{J})$ for sufficiently small $\zeta$, a contradiction.

We, therefore, conclude that there is a $\beta^*$ such that w.h.p, for any sub-instance $\mathcal{J}$ with $\alpha^* n/2 \leq |\mathrm{var}(\mathcal{J})| \leq \alpha^*$, $|\mathcal{B}_M(\mathcal{J})| \geq \beta^* n$, i.e., formula (A.10) holds. □

## A.3 Upper Bound on the Threshold of the Random Restricted Binary CSP $\mathcal{B}_{n,m}^{d,t}[\mathcal{L}]$

In this subsection, we show that the condition mentioned at the end of Definition 2.1 guarantees the existence of a constant $c^*$ such that $\mathcal{B}_{n,m}^{d,t}[\mathcal{L}]$ with $m/n > c^*$ is asymptotically unsatisfiable with probability one. We state and prove the result for the case of binary constraints, but similar results also hold for the more general $k$-ary constraints.

Recall that the condition on the set of relations $\mathcal{L} = \{L_1, L_2, \ldots \mid L_i \subset D \times D\}$ is

$$(a, a) \notin \bigcap_{i \geq 1} L_i, \forall a \in D. \tag{A.13}$$

Since we assume that the domain size is fixed, for any $a \in D$ the probability that a constraint in $\mathcal{B}_{n,m}^{d,t}[\mathcal{L}]$ contains $(a, a)$ as one of its nogoods is lower bounded by a constant $p$.

**Theorem A.5.** *If $\mathcal{B}_{n,m}^{d,t}[\mathcal{L}]$ is such that the set of relations $\mathcal{L}$ satisfies the condition A.13, then it is unsatisfiable w.h.p. if $\frac{m}{n} = n$ is such that*

$$d(1 - \frac{p}{d^2})^c < 1.$$

*Proof.* For any assignment $A$, there must be a subset $S$ of $\frac{n}{d}$ variables that are assigned a common domain value $a$. Then, $A$ satisfies $\mathcal{B}_{n,m}^{d,t}[\mathcal{L}]$ only if $A$ satisfies all the constraints that only involve the variables in $S$. Let $H_i$ be the event that $\mathcal{B}_{n,m}^{d,t}[\mathcal{L}]$ has $i$ constraints that only involve the variables in $S$. Then, we have

$$P\{A \text{ satisfies } \mathcal{B}_{n,m}^{d,t}[\mathcal{L}]\} \leq \sum_{i=0}^{cn} (1-p)^i P\{H_i\}.$$

Let $N = \frac{1}{2}n(n-1)$. Write

$$N_1 = \frac{1}{2}\frac{n}{d}(\frac{n}{d} - 1) = \frac{1}{2d^2}n(n-1) + O(n)$$

for the number of possible edges induced by $S$, and

$$N_2 = N - N_1 = \frac{1}{2}(1 - \frac{1}{d^2})n(n-1) + O(n).$$

We have for any $1 \leq i \leq cn - 1$

$$
\begin{aligned}
P\{H_i\} &= \frac{\binom{N_1}{i}\binom{N_2}{cn-i}}{\binom{N}{cn}} \\
&= \frac{N_1(N_1-1)\cdots(N_1-i+1) * N_2(N_2-1)\cdots(N_2-cn+i+1)}{N(N-1)\cdots(N-cn+1)} \frac{(cn)!}{i!(cn-i)!} \\
&\leq \left(\frac{N_1}{N}\right)^i \left(\frac{N_2}{N-i}\right)^{cn-i} \binom{cn}{i} \\
&\leq \left(\frac{N_1}{N}\right)^i \left(\frac{N_2}{N-cn}\right)^{cn-i} \binom{cn}{i}
\end{aligned}
$$

Similarly, we have

$$P\{H_0\} = \frac{\binom{N_2}{cn}}{\binom{N}{cn}} \leq \left(\frac{N_2}{N-cn}\right)^{cn}$$

and

$$P\{H_{cn}\} = \frac{\binom{N_1}{cn}}{\binom{N}{cn}} \leq \left(\frac{N_1}{N}\right)^{cn}$$

So,

$$P\{A \text{ satisfies } \mathcal{B}_{n,m}^{d,t}[\mathcal{L}]\} \leq \sum_{i=0}^{cn} (1-p)^i P\{H_i\}$$

$$\leq \sum_{i=0}^{cn} (1-p)^i \left(\frac{N_1}{N}\right)^i \left(\frac{N_2}{N-cn}\right)^{cn-i} \binom{cn}{i}$$

$$\leq \sum_{i=0}^{cn} \left((1-p)\frac{N_1}{N}\right)^i \left(\frac{N_2}{N}\frac{N}{N-cn}\right)^{cn-i} \binom{cn}{i}$$

$$= \left((1-p)\frac{N_1}{N} + \frac{N_2}{N}\frac{N}{N-cn}\right)^{cn}$$

$$= \left((1-p)\frac{1}{d^2} + (1-\frac{1}{d^2})\right)^{cn} O(1)$$

$$= \left(1 - \frac{p}{d^2}\right)^{cn} O(1)$$

Therefore,

$$P\{\mathcal{B}_{n,m}^{d,t}[\mathcal{L}] \text{ is satisfiable }\} \leq d^n \left(1 - \frac{p}{d^2}\right)^{cn} = \left(d(1-\frac{p}{d^2})^c\right)^n$$

$\square$